\documentclass[sigconf]{acmart}
\usepackage{graphicx}
\usepackage{float}
\usepackage{xltabular}
\usepackage{booktabs} 
\usepackage{multirow} 
\usepackage{tabularx,array}
\usepackage{makecell}
\usepackage{fvextra}
\usepackage{xcolor} 
\usepackage{xspace}
\usepackage{algorithm}
\usepackage{algorithmic}
\usepackage[most]{tcolorbox}
\usepackage{enumitem}
\usepackage{listings}
\usepackage{balance}

\newcommand{\model}{\texttt{DoctorAgents}\xspace}
\newcommand{\hint}[1]{\hfill$\triangleright$~#1}

\usepackage{amsmath}
\usepackage{mathtools}
\usepackage{amsthm}
\usepackage{placeins}
\usepackage{makecell}
\usepackage[capitalize,noabbrev]{cleveref}

\theoremstyle{plain}

\theoremstyle{definition}

\theoremstyle{remark}

\renewcommand\footnotetextcopyrightpermission[1]{}

\begin{document}

\title{DoctorAgents: Iterative Agentic Refinement for Small Clinical Temporal Data}

\author{%
\textbf{
Ruilin Wang\textsuperscript{1,2},
Bo-Hong Wang\textsuperscript{1,2},
Elizabeth Kourbatski\textsuperscript{1,2},
Jun Bai\textsuperscript{1,2},
Hegang Chen\textsuperscript{1,2},
Ziyang Song\textsuperscript{1,2},
Gilles Boire\textsuperscript{4,5},
Marie Hudson\textsuperscript{3,*},
Yue Li\textsuperscript{1,2,*}
}
}

\affiliation{%
  \institution{%
    \textsuperscript{1}School of Computer Science, McGill University,
    Montreal, Canada\\
    \textsuperscript{2}Mila -- Quebec AI Institute,
    Montreal, Canada\\
    \textsuperscript{3}Division of Rheumatology, Department of Medicine, McGill University, Montreal, Canada\\
    \textsuperscript{4}Centre intégré universitaire de santé et de services sociaux de l’Estrie – Centre hospitalier universitaire de Sherbrooke (CIUSSSE-CHUS), Sherbrooke, Canada\\
    \textsuperscript{5}Division of Rheumatology, Faculty of Medicine and Health Sciences, University of Sherbrooke, Sherbrooke, Canada\\
  }
   \country{\unskip}
}

\begin{abstract}
Clinical machine learning (ML) has the potential to support high-stakes medical decision-making, but reliable deployment is often constrained by scarce, heterogeneous, and temporal complexity. Developing effective ML pipelines for such data remains time-consuming and error-prone, while existing automated machine learning (AutoML) systems only partially address this challenge because they largely rely on brute-force search over predefined spaces and lack explicit reasoning and memory. We therefore reformulate AutoML for small clinical data from exhaustive search to reasoning-driven refinement.
We propose \model, an agentic AI framework that autonomously constructs and optimizes end-to-end ML pipelines through specialized large language model (LLM) agents for generation, validation, and refinement. \model backpropagates natural-language feedback through textual gradient descent to perform targeted updates without exhaustive search. Experiments across diverse clinical tasks show that \model consistently outperforms established AutoML baselines while producing more interpretable task-specific representations.
\end{abstract}

\keywords{Agentic AI, automated machine learning, small clinical data, LLMs}

\maketitle
\pagestyle{plain}
\begingroup
\renewcommand{\thefootnote}{*}
\footnotetext{Corresponding authors: yueli@cs.mcgill.ca and marie.hudson@mcgill.ca}
\endgroup

\section{Introduction}

Existing clinical prediction tasks are often built from small cohorts, such as rare-disease registries, single-center studies, and narrowly defined treatment populations \citep{van2019sample,riley2019minimum}. Unlike standard tabular benchmarks, these datasets are often high-dimensional, sparse, and temporally irregular. More specifically, temporal clinical records are not simply static tables augmented with timestamps but asynchronous and incomplete observations of an evolving patient state where patients may be observed at different frequencies and over substantially different durations, such that the available temporal resolution varies across individuals and variables \citep{horn2020set,shukla2021multi}. The same clinical value may therefore carry different predictive meaning depending on its observation time, recency relative to the prediction time, and the trajectory preceding it.

Moreover, the observed record reflects both the underlying disease process and the clinical observation process, making it difficult to distinguish true physiological change from variation induced by monitoring and care delivery. Naive aggregation into static patient vectors may fail to preserve temporal ordering, rates of change, persistence, and duration-related signals that are central to clinical progression \citep{xie2022deep}. Conversely, modeling the raw sequence requires careful alignment to task-specific index times and observation windows to avoid incorporating information unavailable at prediction time \citep{shukla2019interpolation}. Clinical records are also shaped by care processes: measurement timing, frequency, recency, and missingness may reflect both patient state and clinical attention \citep{che2018recurrent}.

Reliable prediction therefore requires more than selecting a classifier. Because the relevant temporal structure varies with the clinical endpoint, observation window, and prediction horizon, effective pipelines require leakage-safe, task-specific patient representations that capture temporal changes, variability, recency, and missingness \citep{borisov2022deep}. Constructing such robust end-to-end pipelines remains labor-intensive and requires substantial statistical, computational, and clinical expertise \citep{collins2024tripod+}.

To alleviate this burden, \emph{automated ML (AutoML)} has emerged to automate pipeline construction, including data preprocessing, model selection, and hyperparameter searching. Early systems typically relied on predefined search spaces and black-box optimization, including  Bayesian optimization, evolutionary strategies, and bandit-based methods \citep{hutter2011sequential,feurer2015efficient,thornton2013auto, golovin2017google}. Although empirically effective, they often require substantial compute, provide limited insight into \emph{why} a pipeline works, and remain less suitable for small datasets, rapid prototyping, or domain-specific tasks with subtle statistical pitfalls.

Recent advances in \emph{large language models (LLMs)} have enabled AutoML systems that synthesize executable ML pipelines from natural language specifications \citep{wang2024executable,brown2020language,openai2024gpt4}. Beyond static code generation, LLMs can perform data analysis, scientific reasoning, tool use, and iterative self-revision, allowing them to act as autonomous agents \citep{chen2021evaluating,li2022competition,wang2023self}. Recent work on \emph{agentic AI} further shows that decomposing complex tasks into collaborative roles, such as planning, execution, critique, and optimization, improves reasoning depth and efficiency \citep{wang2024survey,xi2025rise,acharya2025agentic,park2023generative,madaan2023self}.
These advances have motivated LLMs as autonomous ML engineers. For instance, AutoML-Agent \citep{trirat2025automlagent} and ERA \citep{era_aygun2026ai} generate, execute, and iteratively refine ML pipelines toward user-defined objectives. Related works also explored tool-augmented LLMs for data science workflows and notebook automation \citep{zhou2023lima,tang2023toolformer}, including systems such as SciToolAgent \citep{ding2025scitoolagent}, which orchestrates domain-specific tools via a curated scientific knowledge graph but focuses on tool selection rather than end-to-end ML pipeline optimization.

Despite these advances, existing LLM-based AutoML systems remain constrained by \emph{trial-and-error} regime, limited persistent memory, weak reasoning over data properties such as leakage risk and evaluation validity, and reliance on wholesale regeneration rather than \emph{targeted, incremental refinement}. These limitations lead to unstable optimization, inefficient compute use, repetitive errors, and shallow exploration, echoing broader challenges in agentic systems with limited structured memory and feedback propagation \citep{shinn2023reflexion,yao2023react}. A more deliberate paradigm is therefore needed, treating pipeline development as a stateful, reasoning-driven process rather than a sequence of independent generations.

We introduce \model, a multi-agent framework for autonomous ML pipeline optimization that replaces brute-force search with reasoning-driven iteration. Briefly, \model employs a team of specialized LLM agents to analyze data, generate and validate executable pipelines, analyze failures, and refine solutions using structured memory of prior changes and outcomes. As a result, the generated ML pipeline are grounded in feature semantics and rigor. During the iterative pipeline refinement, we use adapted Textual Gradient Descent (TGD) \citep{yuksekgonul2025optimizing} to convert natural-language feedback into localized code updates without regenerating entire pipelines. We further devise two sets of domain-specialized agents that decompose optimization into dedicated data preprocessing and model development agents, enabling deeper exploration of task-specific temporal representations while preserving explicit, editable code. Beyond predictive performance, we assess generated features for technical validity, semantic consistency, temporal appropriateness, and clinical justification. Across diverse clinical tasks, \model achieves stronger performance, stability, robustness, and feature quality than existing AutoML and agentic baselines. In summary, our contributions include
\begin{itemize}[noitemsep, topsep=0pt, leftmargin=1em]
  \item a reasoning-driven agentic AutoML framework for small clinical temporal datasets, combining dataset reasoning, execution validation, memory, and refinement,
  \item a memory-aware optimization that enables long-term learning and prevents repeated failures,
  \item a design of domain-specialized data preprocessing for irregularly observed temporal clinical data, enabling richer task-specific patient representations,
  \item automatically composing clinically meaningful variable from temporal data, and
  \item extensive empirical validation using MIMIC-IV and in-house clinical data, demonstrating robustness, efficiency, and interpretability beyond existing methods.
\end{itemize}

\section{Related Work}

\paragraph{Temporal Clinical Prediction Methods.}

Existing methods for temporal clinical prediction typically rely on standardized preprocessing pipelines or task-specific temporal architectures. MIMIC-Extract \citep{wang2020mimic} supports reproducible ICU time-series preprocessing, while ML models such as DeepCare \citep{pham2016deepcare} and TrajGPT \citep{song2025trajgpt} encode visit importance, irregular intervals, and sparse observations through specialized LSTM or transformer-attention layers, respectively. Although effective, these methods usually assume a fixed representation strategy or model family, leaving limited flexibility to adapt preprocessing, feature construction, and model selection jointly across heterogeneous small clinical datasets.

\paragraph{Foundation Models for Small Tabular Data.}

TabPFN \citep{TabPFN} is a pretrained foundation model that performs well on small tabular classification problems with minimal task-specific tuning. However, its monolithic prediction interface provides limited control over preprocessing, feature construction, evaluation design, and task-specific constraints. This limits its suitability for small clinical datasets, where leakage prevention, temporal aggregation, missingness handling, and interpretable patient-level representations often require explicit pipeline decisions.

\paragraph{LLM-Based AutoML Agents.}

AutoML-Agent \citep{trirat2025automlagent} generates and iteratively improves pipelines using LLMs, while ERA \citep{era_aygun2026ai} combines LLM generation with tree-based search to explore a broader solution space. However, these systems still rely on repeated pipeline generation, with limited persistent memory and weak propagation of evaluation feedback into localized code changes. As a result, optimization can remain unstable and sample-inefficient under limited budgets, especially for small clinical datasets.

\section{DoctorAgents}

\begin{figure*}[t]
    \centering
    \includegraphics[width=\textwidth]{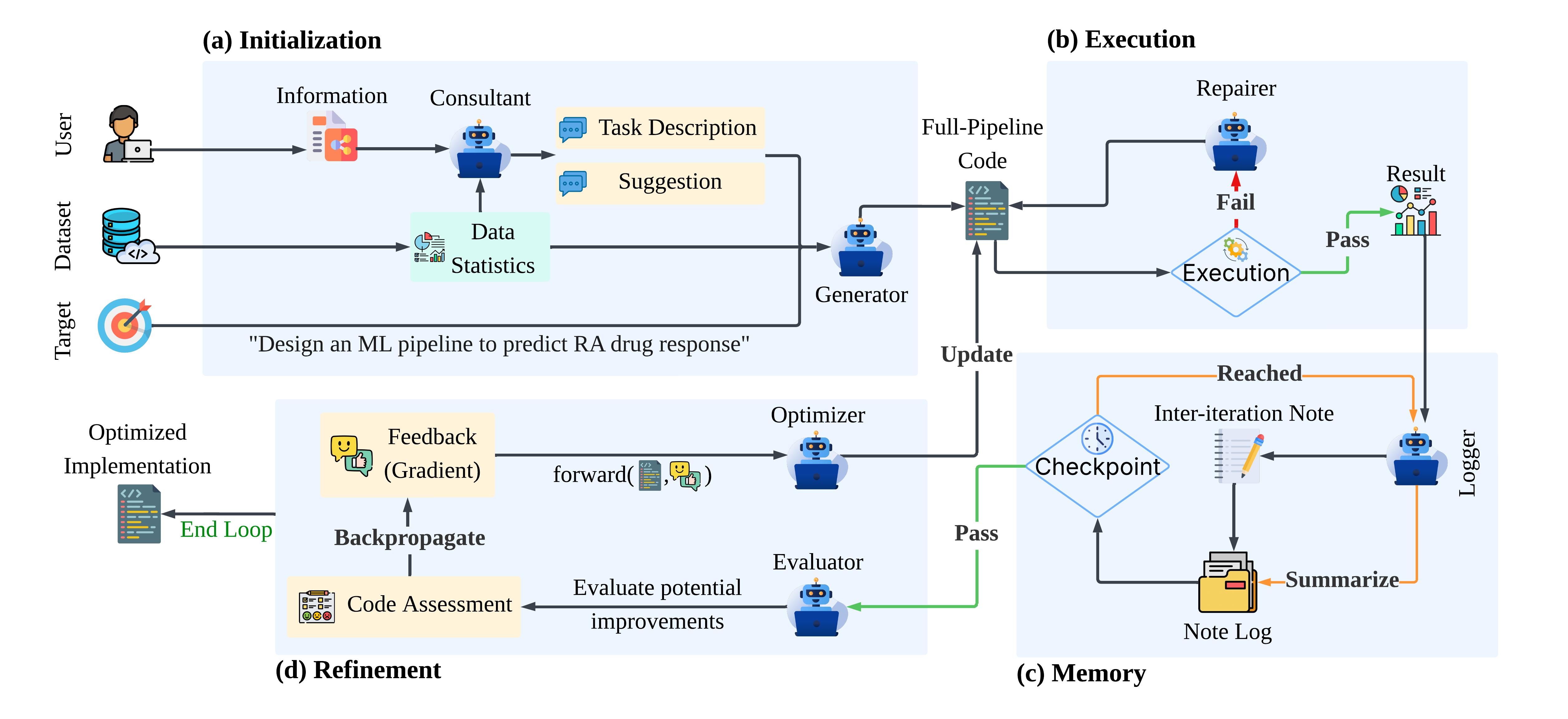}
    \caption{Overview of \model. \textbf{(a) Initialization} produces a solid starting pipeline from user metadata and profiled data statistics using the provided tool environment. \textbf{(b) Execution} validates the given code and returns performance results. \textbf{(c) Memory} records long-term information across optimization iterations. \textbf{(d) Refinement} iteratively improves the pipeline through evaluation-driven updates. The final output to the user is an optimized pipeline with the best observed performance.}
    \Description{\model framework diagram}
    \label{fig:overview}
\end{figure*}

We first describe the overall multi-agent framework of \model with details on how the designated agents handle task initialization, validation, memory log, and pipeline refinement (Fig.~\ref{fig:overview}). We then describe the domain specialized (DS) agents in Sec.~\ref{sec:specialized}.


\subsection{Agent Designs}
Each agent is an LLM instructed to complete a specific task:

\noindentparagraph{Consultant ($\mathbb{A}_c$)}
acts as the high-level analytical expert, responsible for analyzing and transforming raw inputs into a predefined structured task specification with domain knowledge to guide downstream pipeline synthesis. 

\noindentparagraph{Generator ($\mathbb{A}_g$)}
translates the Consultant's outputs into an initial ML pipeline code, including data preprocessing, model selection, training, and evaluation. The resulting implementation forms the starting point for the optimization workflow.

\noindentparagraph{Repairer ($\mathbb{A}_r$)}
restores executability when the current pipeline fails and ensures minimal correction on the code for re-validation until success before proceeding to the Logger.

\noindentparagraph{Logger ($\mathbb{A}_l$)}
records long-term memory for the refinement process. The Logger records inter-step notes that capture meaningful changes and manages memory context through summarization and cleaning at every checkpoint to maintain effective memory log.

\noindentparagraph{Evaluator ($\mathbb{A}_e$)}
evaluates the quality of the pipeline and proposes  improvement strategies. It produces natural language evaluation, which serve as the basis for updating the pipeline in the downstream backprop-optimization process.

\noindentparagraph{Optimizer ($\mathbb{A}_o$)}
converts the backpropagated evaluator feedback into concrete code modifications while preserving expected pipeline output format to maintain iterative stability. The resulting pipeline implementation is then forwarded to the next optimization loop.

\begin{algorithm}
\caption{Overall Workflow of \model}
\label{alg:sk-agent-workflow}

\noindent
\begin{minipage}{\linewidth}
\raggedright
\textbf{Initialization:} Consultant $\mathbb{A}_c$, Generator $\mathbb{A}_g$, Logger $\mathbb{A}_l$, Evaluator $\mathbb{A}_e$, Optimizer $\mathbb{A}_o$, and system step $S$ \\
\textbf{Input:} Dataset metadata $M$ and max optimization step $T$ \\
\textbf{Output:} Optimized Pipeline $P$
\end{minipage}

\begin{algorithmic}[1]

\IF{$M \neq \varnothing$}
    \STATE $I \leftarrow \mathbb{A}_c(M)$ \hint{parse raw inputs}
    \STATE $P^* \leftarrow \mathbb{A}_g(I, M)$ \hint{generate initial code}
    \STATE $R^* \leftarrow \text{Exec}(P^*)$ \hint{validate $P^*$ (Alg.~\ref{alg:execution-validation})}
    \STATE $L \leftarrow \mathbb{A}_l(R^*)$ \hint{initialize memory log}
\ELSE
\STATE \textbf{return} $msg$ \hint{error msg for invalid inputs}
\ENDIF

\WHILE{$S \leq T$}
    \STATE $E \leftarrow \mathbb{A}_e(P^*, M, L)$ \hint{evaluate $P^*$ (Sec.~\ref{sec:optimization})}
    \STATE $F \leftarrow \nabla_{P^*} (E)$ \hint{gradient computation (Eq.~\ref{eq:gradient})}
    \STATE $P \leftarrow \mathbb{A}_o(P^*, F)$ \hint{update $P^*$ (Eq.~\ref{eq:optimize})}
    \STATE $R \leftarrow \text{Exec}(P)$
    \STATE $note \leftarrow \mathbb{A}_l(P^*, P, R^*, R)$ \hint{create note (Sec.~\ref{sec:notebook})}
    \STATE $L \leftarrow L + note$
    \IF{$CP(L)$ = \textsc{true}} 
        \STATE $L \leftarrow \text{SUMMA}(\mathbb{A}_l, L)$ \hint{summarize L (Sec.~\ref{sec:notebook})}
    \ENDIF
    \STATE $P^* \leftarrow P$
    \STATE $R^* \leftarrow R$
    \STATE $S \leftarrow S + 1$
\ENDWHILE

\STATE \textbf{return} $P$

\end{algorithmic}
\end{algorithm}

\subsection{Overall workflow}
The workflow of \model is depicted in Fig.~\ref{fig:overview} and Alg.~\ref{alg:sk-agent-workflow}, which consists of two main stages:

\paragraph{Initialization} As illustrated in Fig.~\ref{fig:overview}a, \textbf{Consultant} ($\mathbb{A}_c$) is prompted to receive validated user-provided information and metadata of the dataset (Line~2) and transforms them into a structured task specification with explicit requirements. These data are then passed to \textbf{Generator} ($\mathbb{A}_g$), which is prompted to produce an initial end-to-end full-pipeline implementation followed by validation (Lines~3--4). The resulting data are added to the memory log for subsequent downstream optimization (Line~5).
    
\paragraph{Optimization} The framework iteratively refines the pipeline to search for an improved implementation by cycling through 3 steps. 
\underline{Step 1 - code execution} (Fig.~\ref{fig:overview}b): \textbf{Repairer} ($\mathbb{A}_r$) receives error traces upon execution failure and attempts to minimally fix the code before re-validating it (Alg.~\ref{alg:execution-validation}). Successful execution returns performance results (Lines~4, 13). 
\underline{Step 2 - memory composition} (Fig.~\ref{fig:overview}c): given both the previous and updated pipelines together with the execution results, \textbf{Logger} ($\mathbb{A}_l$) appends notes that capture important changes to the log (Lines~14--15). Log summarization and cleaning are periodically triggered at predefined checkpoints to minimize the in-context length and memory storage (Line~17). 
\underline{Step 3 - pipeline refinement} (Fig.~\ref{fig:overview}d): \textbf{Evaluator} ($\mathbb{A}_e$) assesses the pipeline using both the latest performance records and the history log (Line~10). Its feedback is backpropagated to \textbf{Optimizer} ($\mathbb{A}_o$) through TextGrad \cite{yuksekgonul2025optimizing}, which updates the pipeline accordingly (Lines~11--12). The updated pipeline is then validated for the next optimization iteration (Line~19) and returned to the user once the stopping criterion is met.

The following subsections provide details of the key steps including Pipeline Initialization (Sec.~\ref{sec:initialization}), Implementation Verification (Sec.~\ref{sec:execution}), Memory Composition (Sec.~\ref{sec:notebook}), Pipeline Refinement (Sec.~\ref{sec:optimization}), and Domain-Specific Specialization (Sec.~\ref{sec:specialized}).

\subsection{Pipeline Initialization}

\label{sec:initialization}

The system begins with an initialization phase to ensure a valid base pipeline (Fig.~\ref{fig:overview}a). As the performance of an autonomous system is highly sensitive to its starting point, a good initialization can achieve the optimal performance at substantially fewer iterations.

\paragraph{Metadata}
To enhance data exploration and improve downstream generation quality, we derive a set of descriptive statistics from the input dataset $D$. Specifically, a small subset of representative samples, feature statistics (computed by LLM-based profiling if not provided), and user-provided background information of the task are combined into a unified metadata representation $M$, which is provided to both $\mathbb{A}_c$ and $\mathbb{A}_g$ to support robust, data-informed pipeline initialization and task understanding.

\paragraph{Task Specification}
The Consultant $\mathbb{A}_c$ takes as input the metadata $M$ and produces a structured output $I$ = $\mathbb{A}_c$($M$) as strictly predefined in the prompt (Sec.~\ref{prm:cons}), consisting of two parts (Example output in Sec.~\ref{example:cons}):
\begin{itemize}[noitemsep, topsep=0pt]
    \item \textbf{Task Description}: a concise and unambiguous description of the target ML task, intended to serve as a stable reference for downstream agents.
    \item \textbf{Task Suggestion}: task-specific recommendations for data preprocessing and model search strategies, including candidate algorithms, normalization schemes, and missing-value handling, which guide subsequent agent decisions.
\end{itemize}

\paragraph{Pipeline Synthesis}
Conditioned on $I$ and $M$, $\mathbb{A}_g$ synthesizes the initial executable pipeline $P^*$. Specifically, it translates the task description, variable statistics, and reference suggestions into a complete pipeline implementation, including \textit{data loading}, \textit{preprocessing}, \textit{model construction}, \textit{training}, \textit{evaluation}, and \textit{result reporting}. $\mathbb{A}_g$ is encouraged to follow the suggestion by $\mathbb{A}_c$, but it is not constrained to a fixed template, enabling flexibility across diverse tasks and datasets. The resulting $P^*$ = $\mathbb{A}_g$($I, M$) serves as the initial node in the downstream optimization trajectory.

\subsection{Implementation Verification}
\label{sec:execution}

Given a pipeline code $P_e$, \model validates both the executability and efficiency. This stage integrates program execution with an autonomous repair mechanism (Fig.~\ref{fig:overview}b; Alg.~\ref{alg:execution-validation}), forming a closed-loop safeguard before downstream optimization.

Execution validation (\textbf{Exec}) attempts to execute the pipeline code through a program runner $f_e$ within a fixed time budget and produces an execution signal $flag$, an error message $err$, and results $R_e$. Successful execution ($flag$ = PASS) yields concrete outputs $R_e$, involving \textit{training}, \textit{validation}, and \textit{test performance} of the model with its \textit{name} and \textit{hyperparameter matrix}, or metadata such as \textit{decision-oriented strategies and variable-level information}, which will be subsequently consumed by the downstream agents. Program failures ($flag$ = FAIL) may arise from a variety of sources, including runtime errors, incompatible library usage, or resource exhaustion. When execution fails, the error trace $err$ (either an execution error or a timeout resulting from the time budget being exceeded), together with $P_e$, is passed to $\mathbb{A}_r$, which is prompted (Sec.~\ref{prm:rep}) to minimally modify the program to restore executability of $P_e$ via $\mathbb{A}_r$($P_e$, $err$) while preserving the original algorithmic intent. Typical repair actions include correcting method misuse, resolving variable scope or type errors, and mitigating runtime exceptions. After producing a repaired version of $P_e$, re-validation is performed. This procedure repeats until either the program executes successfully or a predefined retry limit is reached (which is rare). By incorporating execution validation into optimization, \model resolves execution-level issues for successful execution.
 
\subsection{Memory Composition}
\label{sec:notebook}

A key challenge in designing an autonomous code optimization system is the lack of a persistent state across iterations. Without explicit memory, agents may repeatedly revisit identical or similar failures, reintroduce previously corrected errors, or overlook insights obtained from earlier iterations. To address this issue, $\mathbb{A}_l$ maintains a structured, long-term memory throughout the optimization process (Fig.~\ref{fig:overview}c).

\paragraph{Memory Log}
Following pipeline initialization (Sec.~\ref{sec:initialization}), $\mathbb{A}_l$ initializes the memory log $L$ using the initial results $R^*$ obtained from Exec($P^*$). Thereafter, $\mathbb{A}_l$ records, organizes, and maintains inter-iteration information $note$ generated during pipeline refinement. Given the updated pipeline $P$ and results $R$ = Exec($P$), rather than storing raw module outputs, $\mathbb{A}_l$ is prompted (Sec.~\ref{prm:log_note}) to extract meaningful \emph{structured inter-iteration pipeline differences} (e.g., changes in preprocessing logic, model configurations, or hyperparameters) and associate them with performance outcome differences. Hence $note$ = $\mathbb{A}_l$($P^*, P, R^*, R$). These $note$s are appended to $L$ and made accessible to downstream agents as historical context that informs subsequent decisions.

\paragraph{Checkpoint ($CP$) and Summarization (SUMMA)}
As optimization progresses, unbounded accumulation of notes can lead to excessive context length and degrade reasoning quality. To mitigate this \emph{context explosion} problem, $\mathbb{A}_l$ incorporates a checkpoint-based memory management strategy. At the predefined checkpoints such as every 15 optimization iterations or at the 5000-token limit ($CP$($L$) = \textsc{true}), $\mathbb{A}_l$ triggers a cleaning procedure \textbf{SUMMA} that summarizes its memory following the instruction prompts (Sec.~\ref{prm:log_clean}). This involves distilling notes from earlier iterations, discarding low-priority or redundant entries, and retaining only salient insights (e.g., changes in codes that significantly improved or degraded performance; an example shown in Sec.~\ref{example:log}) along with a small set of most recent observations. With this knowledge summarization strategy, the resulting log $L$ = SUMMA($\mathbb{A}_l$, $L$) preserves long-term memory and maintains concise information.

\subsection{Pipeline Refinement}
\label{sec:optimization}
Once the initial pipeline $P^*$ is validated (Sec.~\ref{sec:execution}) to initialize the optimizer and the first $note$, derived from the execution result $R^*$ = Exec$(P^*)$, which is cached into $L$, \model enters the core \emph{Refinement} phase, where pipeline quality is iteratively improved through evaluation-driven program refinement. This phase follows a closed-loop structure consisting of two primary steps: \textbf{Evaluation} and \textbf{Optimization} (Fig.~\ref{fig:overview}d).

\paragraph{Evaluation}
Given the current pipeline $P^*$, dataset metadata $M$, and the memory log $L$, $\mathbb{A}_e$ produces an assessment $E$ that typically covers three aspects (Sec.~\ref{prm:eval}):
\begin{itemize}[noitemsep, topsep=0pt]
    \item \textbf{Correctness}: check if $P^*$ correctly implements the intended task and evaluation protocol, and suggest corrections otherwise;
    \item \textbf{Identified Flaws}: detect weaknesses such as overfitting, inefficient preprocessing, or inappropriate model choices with respect to $L$ and suggest a mitigation plan;
    \item \textbf{Improvement Opportunities}: provide suggestions to improve performance, robustness, and efficiency.
\end{itemize}
Unlike numerical optimization in traditional ML, $E$ = $\mathbb{A}_e$($P^*$, $M$, $L$) is expressed in natural language, capturing fine-grained and context-aware judgments about pipeline behavior and enabling more informative downstream updates.

\paragraph{Optimization via Backpropagating TGD}
The backpropagation component translates $E$ into localized instructions for code modification. Analogous to the numerical gradient backpropagation, it propagates the high-level evaluation signal backward through the \textit{program representation} with respect to $P^*$, identifying components that need to be improved, replaced, or discarded \cite{yuksekgonul2025optimizing}. This produces an actionable feedback signal: 
\begin{equation}\label{eq:gradient}
    F = \nabla_{P^*}(E),
\end{equation}
which $\mathbb{A}_o$ receives as input to update the pipeline as follows:
\begin{equation}\label{eq:optimize}
    P = \mathbb{A}_o(P^*, F),
\end{equation}    
where it incorporates changes to preprocessing plan, model configuration, hyperparameters, training procedures, and/or evaluation protocols. $\mathbb{A}_o$ performs incremental, targeted modifications while preserving previously successful components, avoiding costly full regeneration. 
After optimization, the revised pipeline $P$ is executed to obtain new results $R$ for memory logging, and then both are forwarded back as $P^*$ and $R^*$ to close one optimization iteration. This iterative process continues until the termination criterion is met (e.g., predefined optimization steps $T$).

\begin{figure}
    \centering
    \includegraphics[width=\columnwidth]{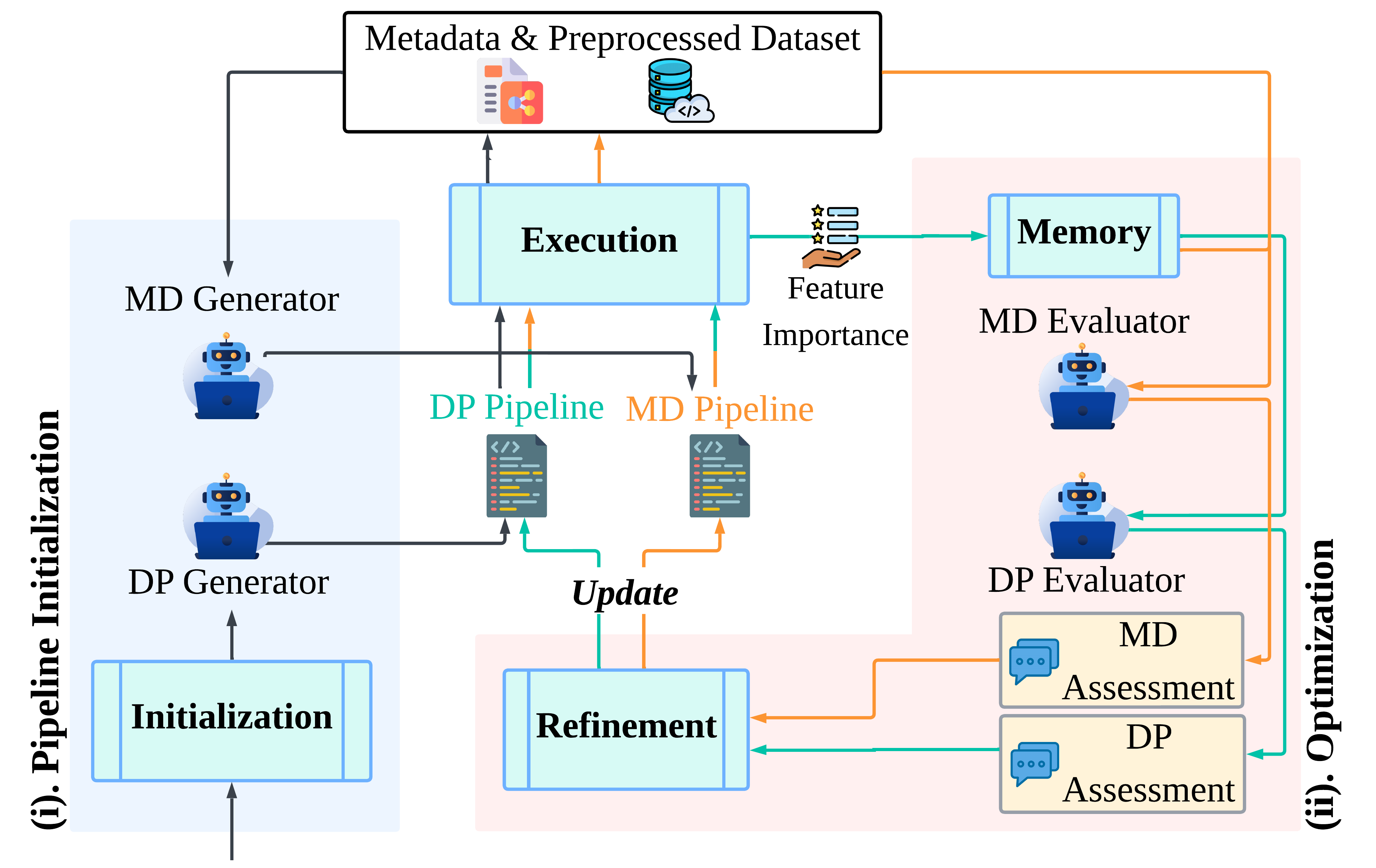}
    \caption{\model Domain-Specific (DS) Design. Both Generator and Evaluator are specialized into Data Preprocessing (DP) and Model Development (MD) to enable deeper DS exploration and targeted implementation.}
    \Description{\model-DS framework diagram.}
    \label{fig:new_framework}
\end{figure}

\subsection{Domain-Specific Specialization}
\label{sec:specialized}

Building upon the \model framework, we introduce a specialized variant that decomposes the pipeline into two domain-specific subtasks—Data Preprocessing (\textbf{DP}) and Model Development (\textbf{MD})—and assigns them to dedicated agents as Generator agents, $\mathbb{A}_g^{dp}$ and $\mathbb{A}_g^{md}$, and Evaluator agents, $\mathbb{A}_e^{dp}$ and $\mathbb{A}_e^{md}$ (Fig.~\ref{fig:new_framework}). This decomposition allows agents to operate within focused search domains: DP Agents focus on identifying feature engineering strategies, while MD Agents explore a richer spectrum of model families and training configurations. 

\paragraph{Initialization}
$\mathbb{A}_g^{dp}$ first generates a preprocessing-only  pipeline $P_{dp}^* = \mathbb{A}_g^{dp}(I, M)$ (Sec.~\ref{prm:dp_gen}) given the task specification $I$ and metadata of the dataset $M$. Intermediate data $R_i^*$ are produced via Exec$(P_{dp}^*)$, including enriched feature representations, metadata describing applied transformations, and a preprocessed dataset, thereby supporting broader exploratory behavior in the preprocessing domain. The data $R_i^*$ are then forwarded to $\mathbb{A}_g^{md}$ to synthesize the model-development pipeline $P_{md}^* = \mathbb{A}_g^{md}(I, R_i^*)$ (Sec.~\ref{prm:mt_gen}). Executing $P_{md}^*$ produces results $R^*$ which include information about top features ranked by importance (computed with SHapley Additive exPlanations (SHAP) \cite{shap_lundberg2017unified}) and the initial memory log $L = \mathbb{A}_l(R_i^*, R^*)$. 

\paragraph{Optimization} 
$\mathbb{A}_e^{dp}$ evaluates $P_{dp}^*$ (Sec.~\ref{prm:dp_eval}) using $M$ and $L$ ($E_{dp} = \mathbb{A}_e^{dp}(P_{dp}^*, M, L)$, an example shown in Sec.~\ref{example:eval}) to backpropagate domain-specific feedback $F_{dp} = \nabla_{P_{dp}^*}(E_{dp})$ (e.g., Sec.~\ref{example:dp_fb}), following an evaluation scheme tailored to the data preprocessing domain. The updated pipeline is then produced as $P_{dp} = \mathbb{A}_o(P_{dp}^*, F_{dp})$. Executing $P_{dp}$ yields new intermediate data $R_i = \text{Exec}(P_{dp})$, which are passed to generate assessment $E_{md} = \mathbb{A}_e^{md}(P_{md}^*, R_i, L)$ (prompt described in Sec.~\ref{prm:mt_eval}) for $P_{md}^*$ (e.g., Sec.~\ref{example:mt_eval}). The resulting feedback $F_{md} = \nabla_{P_{md}^*}(E_{md})$ (e.g., Sec.~\ref{example:mt_fb}) produces refined $P_{md}$ via $\mathbb{A}_o(P_{md}^*, F_{md})$. Memory log is computed between original and updated pipelines ($P^*$s and $P$s) along with the execution results after optimizing $P_{md}^*$ by Logger $\mathbb{A}_l$, and then cached to $L$ for future evaluations. Executing $P_{md}$ produces updated performance results $R$ for the current iteration. In the end, $P_{dp}, P_{md}, R_i, R$ are forwarded back for the next iteration. Together, the DP and MD updates form one complete optimization loop.

\section{Experiments}
\subsection{Benchmark Datasets}
We evaluated \model on four temporal clinical datasets (Table~\ref{tab:task_setup}), including three tasks derived from the public MIMIC-IV database \citep{mimic4} and one rheumatoid arthritis (RA) dataset from the Early Undifferentiated PolyArthritis (EUPA) cohort \citep{carrier2025changes, carrier2026longitudinal}. The MIMIC-IV tasks are: (1) ICU mortality prediction, which predicts in-ICU mortality using the first 48 hours of data; (2) ICU readmission prediction, which predicts readmission within seven days after the first ICU discharge using all pre-discharge observations; and (3) ICU length-of-stay (LOS) prediction, which predicts whether an ICU stay exceeds 3 days using the first 24 hours of data. These datasets include irregular temporal laboratory measurements from \texttt{labevents} and selected vital signs from \texttt{chartevents}.

The EUPA dataset contains 1--9 years of sparsely sampled longitudinal rheumatoid arthritis (RA) patient records and is used to predict anti-TNF treatment response from pre-treatment clinical trajectories, distinguishing responders from non-responders. 

For all datasets, temporal observations were retained with their raw irregular timestamps within each task-specific observation window, without fixed binning or manual visit selection. This design preserves the original temporal structure and allows \model to determine suitable aggregation, recency, and change-based representations during pipeline construction.

We adopted task-specific evaluation metrics according to the class distributions. We report the area under the receiver operating characteristic curve (AUROC) for the approximately balanced RA treatment-response prediction task and the area under the precision--recall curve (AUPRC) for the three MIMIC-IV tasks, for which the outcome labels are imbalanced.

\subsection{Baseline Methods}
We evaluated TabPFN as a traditional ML competitor for small-sample tabular learning \cite{TabPFN}. We also compared against three LLM-based few-shot baselines: a state-of-the-art (SOTA) general-purpose proprietary model GPT-5 \cite{singh2025openaigpt5card}, and two open-source LLMs GPT-oss-20b and Llama-8b \cite{gpt-oss, grattafiori2024llama}. To evaluate SOTA agentic approaches, we benchmarked AutoML-Agent  \cite{trirat2025automlagent} and ERA \cite{era_aygun2026ai}. GPT-5 was used as the backbone LLM for all agentic frameworks including ours. The prompt details are provided in Sec.~\ref{apsec:exp_design}.

\subsection{Evaluation Design}

To ensure rigorous evaluation, each dataset was partitioned into a meta-train and meta-test split under a stratified five-fold cross-validation design. The final results were reported on the meta-test split as the mean and standard deviation across the five folds. In order to evaluate the agents' ability to reason over and leverage dataset information during preprocessing, all datasets were only minimally processed to prevent data leakage and were otherwise kept as close to their raw form as possible. 

For \model, the meta-train split was further divided into train, validation, and internal test subsets, where machine learning models were trained on the train split. Optimization decisions were driven by performance on the validation split, while the best pipeline selection was based on the internal test split. For a fair comparison, in each fold, \model and ERA were limited to a maximum of 20 optimization steps and search nodes, respectively, whereas \model-DS was limited to 15 optimization steps. The retry limit for both \model and AutoML-Agent was set to 5.
Since TabPFN does not support time-series inputs, we applied an additional compatibility-only preprocessing step for TabPFN, representing each feature by its last observed value within the observation window. TabPFN was then directly trained on the entire meta-train split and evaluated on the meta-test split, while all other baselines leveraged the meta-train's train-valid-test partitions to synthesize their ML pipelines.
All experiments were run on an NVIDIA 4x RTX PRO 6000 Blackwell Max-Q workstation.

\begin{figure*}[t!]
    \centering
    \includegraphics[width=\textwidth]{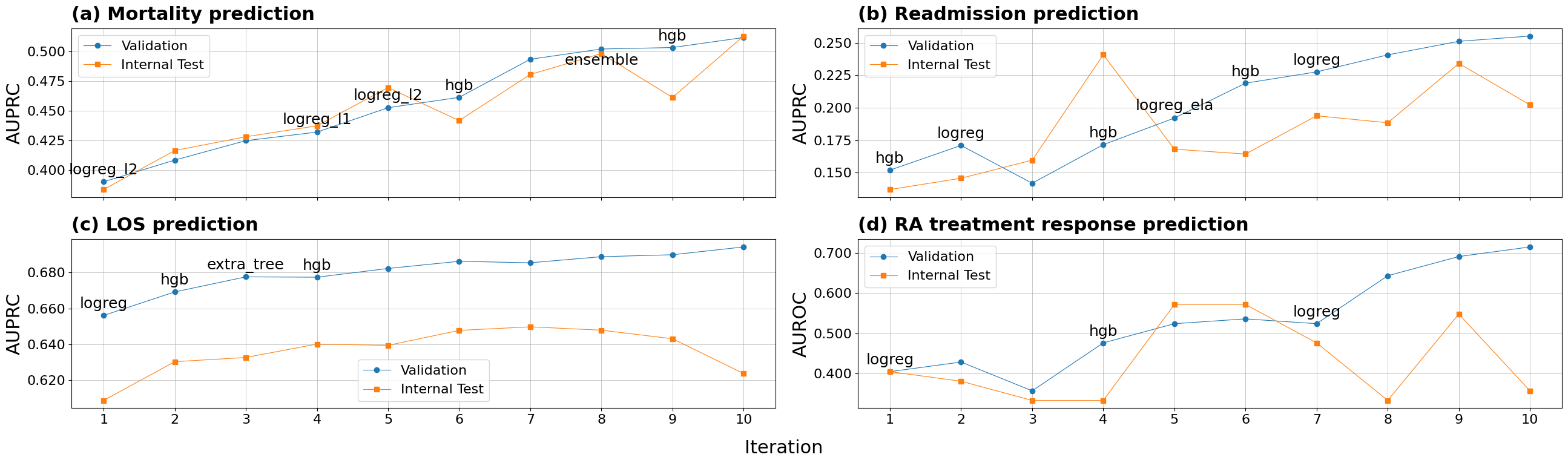}
    \caption{Validation and internal test performance over iteration for \model. Each iteration was labelled with the model name if a new model was selected. The unlabeled iterations indicate that the best model remains the same although there could be preprocessing or hyperparameter updates. For illustration purposes, only 10 iterations are shown although \model can reach higher validation performance for more iterations. A table for model name explanation is provided in Table~\ref{tab:model_exp}.}
    \Description{Iterative improvements curves.}
    \label{fig:improv_curve}
\end{figure*}

\section{Results}

\subsection{Iterative ML Pipeline Optimization}
We first validated the iterative refinement behavior of \model (Fig.~\ref{fig:improv_curve}). Across all benchmarks, validation performance showed a generally improving trajectory, indicating that successive agentic iterations apply consistently beneficial refinements. In parallel, \model dynamically revised its choice of learning algorithms, as reflected by the model annotations on the curve.
For mortality prediction (Fig.~\ref{fig:improv_curve}a), \model started from a regularized linear model and progressively explored alternative configurations, including boosting and ensemble-based variants, while validation AUPRC improved through subsequent preprocessing and hyperparameter refinements. A similar pattern was observed for readmission prediction and RA anti-TNF response prediction (Fig.~\ref{fig:improv_curve}b,d), where the best-performing models across iterations were mainly selected from regularized logistic regression and histogram-based gradient boosting. Notably, \model evaluated a broader candidate model space at each iteration, including XGBoost, Random Forest, LightGBM, and ensemble variants. The repeated selection of linear and boosting-based models suggests that they provide strong inductive biases for high-dimensional, sparse, and imbalanced small clinical datasets. In contrast, LOS prediction (Fig.~\ref{fig:improv_curve}c) involved transitions among linear, boosting, and extra-tree models, indicating that non-linear tree ensembles may better capture interaction patterns associated with prolonged ICU stays.

\begin{table*}[t!]
\centering
\caption{Performance benchmarking on the meta-test data, reported as mean $\pm$ std across 5 cross-validation folds.}
\label{tab:performance_report}
\resizebox{\textwidth}{!}{
\begin{tabular}{lcccc}
\toprule
Method / Task (Metrics) & mortality (AUPRC) & readmission (AUPRC)& LOS (AUPRC) & RA treatment response (AUROC) \\
\midrule
TabPFN \cite{TabPFN} & 
0.509 $\pm$ 0.034 & 0.163 $\pm$ 0.039 & 0.688 $\pm$ 0.022 & 0.539 $\pm$ 0.163 \\
GPT-oss \cite{gpt-oss} & 
0.443 $\pm$ 0.046 & 0.149 $\pm$ 0.033 & 0.675 $\pm$ 0.023 & 0.471 $\pm$ 0.139  \\
Llama-8B \cite{grattafiori2024llama} & 
0.447 $\pm$ 0.035 & 0.142 $\pm$ 0.042 & 0.664 $\pm$ 0.037 & 0.453 $\pm$ 0.106  \\
GPT-5 \cite{singh2025openaigpt5card} & 
0.456 $\pm$ 0.043 & 0.154 $\pm$ 0.020 & 0.678 $\pm$ 0.026 & 0.498 $\pm$ 0.161  \\
AutoML-Agent \cite{trirat2025automlagent} & 
0.475 $\pm$ 0.022 & 0.145 $\pm$ 0.010 & 0.689 $\pm$ 0.020 & 0.520 $\pm$ 0.122  \\
ERA \cite{era_aygun2026ai} & 
0.476 $\pm$ 0.014 & 0.192 $\pm$ 0.044 & 0.692 $\pm$ 0.024 & 0.564 $\pm$ 0.121\\
\midrule
\textbf{\model (ours)} & 
0.482 $\pm$ 0.051 & 0.190 $\pm$ 0.017 & 0.696 $\pm$ 0.028 & 0.505 $\pm$ 0.060 \\
\textbf{\model-DS (ours)}  & 
\textbf{0.520 $\pm$ 0.076} & \textbf{0.212 $\pm$ 0.047} & \textbf{0.704 $\pm$ 0.019} &\textbf{0.577 $\pm$ 0.088} \\
\bottomrule
\end{tabular}
}
\end{table*}

\subsection{Method Comparison}
The domain-specific \model-DS achieves the strongest mean performance across all benchmarks, while \model remains competitive against the established agentic and LLM-based baselines (Table~\ref{tab:performance_report}). For mortality prediction, \model-DS obtains the highest AUPRC, surpassing TabPFN and ERA. 
\model-DS also outperforms all baselines by constructing richer temporal representations for readmission prediction task.
On the prolonged LOS task, both \model variants perform competitively, with \model-DS achieving the highest AUPRC. 
\model-DS also attains the highest AUROC on the small longitudinal RA cohort.
These results demonstrate that the framework adapts its preprocessing and model-selection strategies to heterogeneous prediction horizons, feature distributions, and sample sizes without relying on manually designed task-specific pipelines or modality-specific foundation models. 

\begin{figure}[t]
    \centering
    \includegraphics[width=\columnwidth]{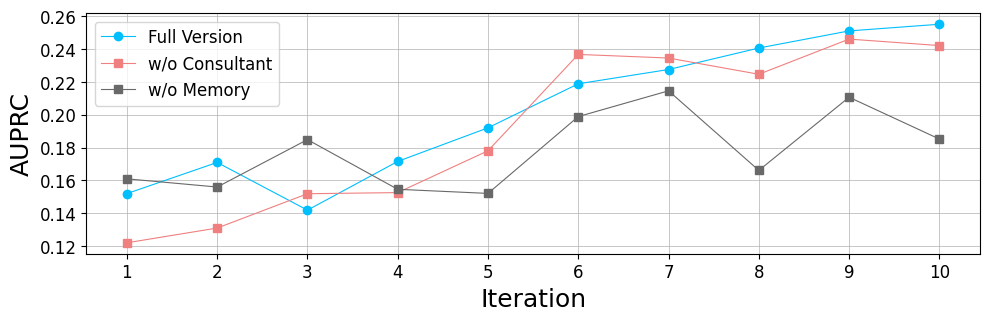}
    \caption{Ablation study on ICU readmission prediction as an example, comparing validation AUPRC across optimization iterations for the full \model and variants without the Consultant or Memory module.}
    \Description{Ablation study.}
    \label{fig:ablation}
\end{figure}

\subsection{Ablation Study}
\label{sec:ablation}

We conducted an ablation study on the contribution of the Consultant and persistent Memory to the iterative optimization process (Fig.~\ref{fig:ablation}), which are designed to improve initialization quality, optimization stability, and long-term refinement over the essential backbone of \model. The full \model exhibits the most consistent overall improvement and achieves the highest validation AUPRC among ten iterations on ICU readmission prediction.

Removing the Consultant primarily weakens pipeline initialization: the corresponding variant started from a lower validation AUPRC and remained below the full system during most early iterations, although subsequent refinement partially closed the gap. This suggests that the Consultant provided a better-informed initial task specification and preprocessing direction, reducing the amount of exploration required in later steps. Without Memory, the optimization trajectory became substantially less stable, with transient improvements followed by pronounced performance degradation. Although \model without Memory achieved a high validation score in the early iterations, it failed to retain and exploit the corresponding successful strategy in subsequent refinements. This behavior indicates that persistent records of previous modifications and outcomes help prevent the optimizer from revisiting ineffective strategies or overwriting previously successful components.

All ablated variants attain a lower maximum validation AUPRC than the complete system. In particular, the full version reached approximately 0.255, compared with 0.246 without the Consultant and 0.215 without Memory. These results demonstrate that these modules play distinct but complementary roles in the optimization process: the Consultant supports effective initialization whereas Memory stabilizes prolonged iterative refinement.

\section{Interpretability}
\subsection{In-code Feature Composition}
\label{sec:code_analysis}
Besides selecting suitable models, \model-DS autonomously constructs task-specific representations measured by mean absolute SHAP values from temporal clinical data (Fig.~\ref{fig:beeswarm}, Table~\ref{tab:features_exp}). 

For mortality prediction, the leading features characterize early ICU physiological instability across the first 48 hours and baseline clinical vulnerability (Fig.~\ref{fig:beeswarm}a, Sec.~\ref{code:mort_example_code}). They include metabolic and acid-base dynamics (\textit{bicarbonate\_blood\_delta\_per\_hr}), neurological status (e.g. \textit{gcs\_eye\_\allowbreak last}), and hemodynamic condition (\textit{mean\_arterial\_\allowbreak pressure\_\allowbreak mean\_\allowbreak 24\_\allowbreak 48h}). The inclusion of \textit{n\_glucose\_\allowbreak blood\_\allowbreak meas\_\allowbreak 48h} further reflects clinical observation intensity. Together, these features align with mortality prediction as an acute deterioration task by combining baseline vulnerability, evolving physiology, and monitoring patterns.

For the readmission task, \model-DS constructs features describing residual instability and clinical attention near ICU discharge (Fig.~\ref{fig:beeswarm}b, Sec.~\ref{code:readm_example_code}), including recent monitoring intensity through heart-rate counts over the last 6 hours (\textit{heart\_\allowbreak rate\_\allowbreak count\_\allowbreak log1p\_\allowbreak last6h}) and short-window physiological trajectories through the 6-hour respiratory-rate slope (\textit{respiratory\_\allowbreak rate\_\allowbreak slope\_\allowbreak last6h\_\allowbreak per\_\allowbreak hr}). These features align with readmission risk, which is tied to unresolved instability immediately before discharge rather than only to early ICU severity.

For prolonged ICU LOS prediction, respiratory support, oxygenation, neurological function, and measurement recency are prominent (Fig.~\ref{fig:beeswarm}c, Sec.~\ref{code:los_example_code}). Features such as \textit{fio2\_mean\_last6h} and \textit{spo2\_fio2\_ratio} summarize oxygenation status and respiratory support requirements. \textit{lactate\_blood\_recency\_hours} distinguishes recent clinical evidence from stale observations, while \textit{gcs}-related features capture neurological status and assessment intensity. This representation matches the LOS task by encoding persistent organ-support needs and delayed recovery signals.

For RA treatment-response prediction, \model-DS converts variable pre-treatment visits into a patient-level representation that includes baseline genetic and serological risk factors, recent disease activity, longitudinal variability and discrete treatment choices (Fig.~\ref{fig:beeswarm}d, Sec.~\ref{code:ra_example_code}). Some of the leading features, including the presence of 2 shared epitope alleles (\textit{SE\_2.0}), anti-CCP antibodies (\textit{antiCCP\_max\_pre}) and fair functional status (\textit{GrFonct\_2.0}), provide baseline indicators of disease severity. Recent disease activity is represented by patient and evaluator global assessments (\textit{PGA\_last\_pre} and \textit{EGA\_last\_pre}). Meanwhile, longitudinal variability in fatigue (\textit{Fatigue\_std\_365d}) and temporal change in platelet count (\textit{FscPLQ\_delta\_365d}) reflect more distant variation in levels of function and inflammation. Finally, the specific type of anti-TNF drug (\textit{AntiTNF\_or\_JAK\_First\_3}, which refers to etanercept as opposed to other anti-TNF drugs) provides treatment level prediction information. The importance of both static genetic and immunological markers and longitudinal pre-treatment changes shows that \model-DS preserves complementary information about patient background, disease trajectory, and current clinical status.

Overall, these results demonstrate that \model-DS composes predictive features according to each task's temporal structure, clinical context, and non-random missing pattern rather than relying on a fixed preprocessing template.

\subsection{Reasoning Validation}

\begin{figure}[t]
    \centering
    \includegraphics[width=\columnwidth]{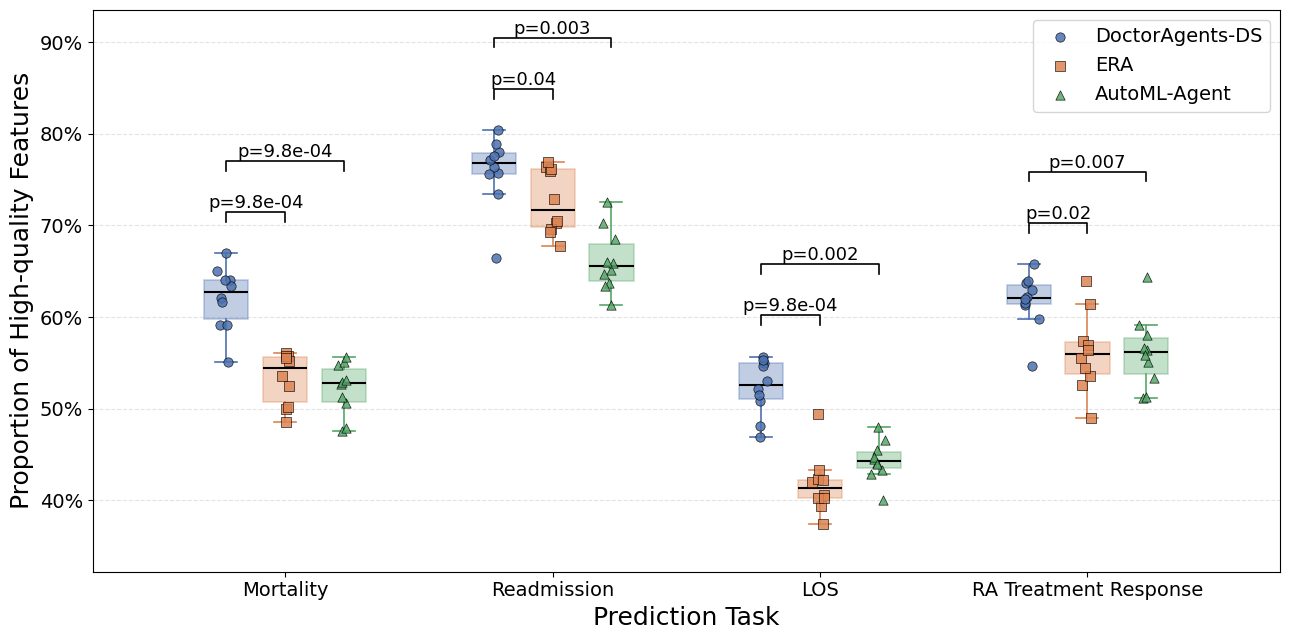}
    \caption{Distribution of high-quality feature proportions composed by \model-DS, ERA, and AutoML-Agent across all tasks. For each method and task, the distribution comprises 10 observations from two independent runs over five folds, each denoting the proportion of \textit{Well Justified} or \textit{Clinically Reasonable} features. P-values were obtained from one-sided paired Wilcoxon signed-rank tests.}
    \Description{High-quality feature distribution boxplot.}
    \label{fig:reasonability}
\end{figure}

To validate the features composed by \model-DS, we assessed the reasonability of features using GPT-5.4. We used GPT-5.4 as an independent judge because of its consistent semantic reasoning over feature names, mathematical operations, and task context with the caveat that GPT-5.4 is not perfect. Features were classified into five levels: \textit{Invalid}, \textit{Weakly Justified}, \textit{Plausible but Uncertain}, \textit{Well Justified}, and \textit{Clinically Reasonable}, based on clinical validity, mathematical correctness, semantic compatibility, and consistency with the task setting (prompt details in Sec.~\ref{instr:feat_reason}).

\model-DS produced the highest proportion of \textit{Well Justified} features across all four tasks (Fig.~\ref{fig:bar_reason}). Although its proportion of \textit{Clinically Reasonable} features was slightly lower than that of ERA in three tasks, this likely reflects its broader use of temporal representations, including changes across observation periods, measurement recency, and visit-level differences. In contrast, ERA and AutoML-Agent relied more heavily on conventional transformations such as last values, means, and missingness indicators, which may be more readily judged as \textit{Clinically Reasonable} as they are widely established in prior work. 

To evaluate overall feature quality, we grouped \textit{Well Justified} and \textit{Clinically Reasonable} features as \textit{high-quality features}. Across the paired evaluations, \model-DS generated a higher overall proportion of high-quality features than both ERA and AutoML-Agent (Fig.~\ref{fig:reasonability}), with mean improvements of $7.10\%$ and $8.33\%$, respectively. One-sided paired Wilcoxon signed-rank tests \citep{wilcoxon1945individual} confirmed that both improvements were significant ($p = 4.61 \times 10^{-8}$ and $8.00 \times 10^{-11}$, respectively).


\section{Discussion}

We developed \model, a reasoning-driven agentic framework for constructing reliable ML pipelines for small clinical datasets. Across all benchmarks, \model achieves competitive performance while adapting to heterogeneous prediction settings. Ablation results show that its components jointly improve initialization and optimization stability, while \model-DS produces interpretable, task-specific representations with stronger performance and significantly more high-quality features than competing agentic methods. Notably, performance on RA treatment response prediction remains modest across methods, likely due to the limited availability of strong predictive biomarkers, with the small cohort further increasing fold-level sensitivity to individual patients. Future work will include extending \model to multimodal data and improving agent specialization through instruction fine-tuning.

\bibliographystyle{ACM-Reference-Format}

\clearpage
\bibliography{main}

\newpage
\appendix

\renewcommand{\thetable}{A\arabic{table}}
\renewcommand{\thefigure}{A\arabic{figure}}
\setcounter{figure}{0}
\setcounter{table}{0}

\lstset{
  basicstyle=\small\ttfamily,
  commentstyle=\upshape,
  breaklines=true,
  breakatwhitespace=false,
  breakautoindent=true,
  breakindent=1em,
  columns=fullflexible,
  keepspaces=true,
  showstringspaces=false,
  tabsize=6
}

\newtcolorbox{sysmsg}[1][]{%
  enhanced,
  breakable,
  colback=gray!5,                
  colframe=blue!50!black,        
  colbacktitle=blue!60!black,    
  coltitle=white,                
  title=#1,
  fonttitle=\bfseries,
  boxed title style={%
    sharp corners,
    boxrule=0pt,
  },
  arc=2mm,                       
  boxrule=0.4pt,
  top=2mm,bottom=2mm,
  left=2mm,right=2mm,
  before skip=6pt,after skip=10pt
}

\clearpage

\fvset{
  breaklines=true,
  breakanywhere=true,
  fontsize=\small,
}


\begin{figure*}[t!]
\section{Appendix}
\subsection{Supplementary Figures}
\label{apsec:beeswarm}
    \centering
    \includegraphics[width=0.45\textwidth]{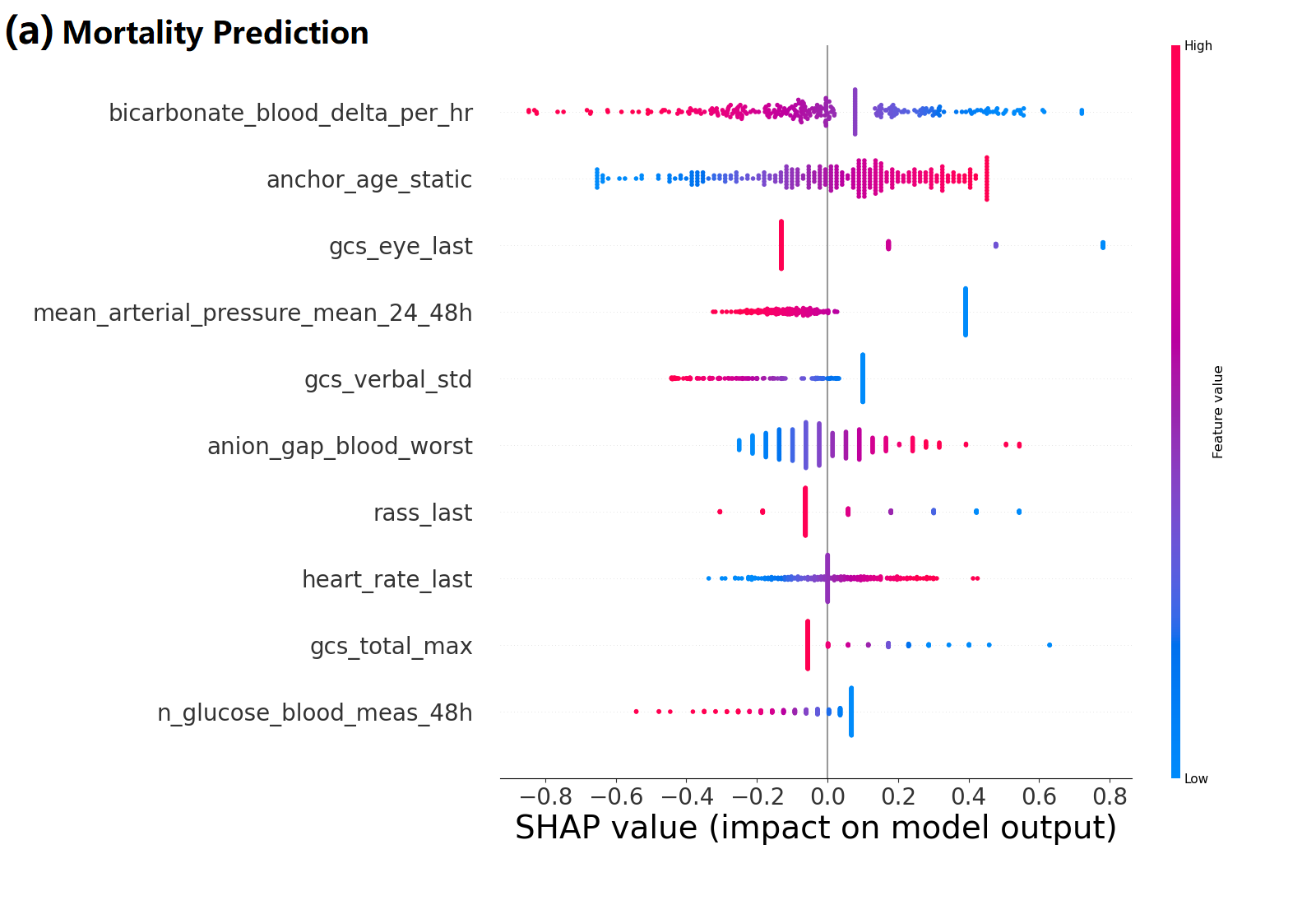}
    \hfill
    \includegraphics[width=0.45\textwidth]{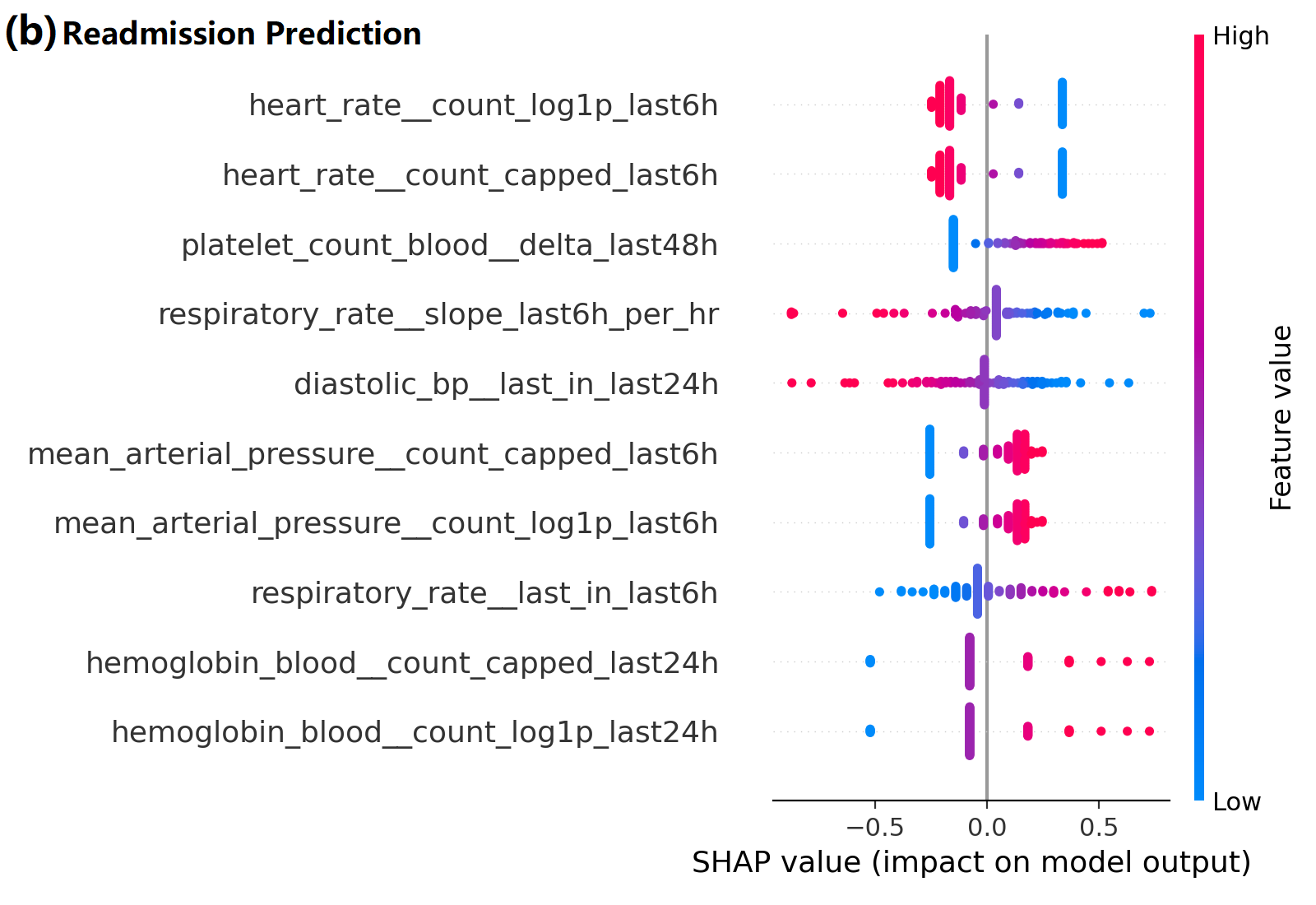}


    \includegraphics[width=0.45\textwidth]{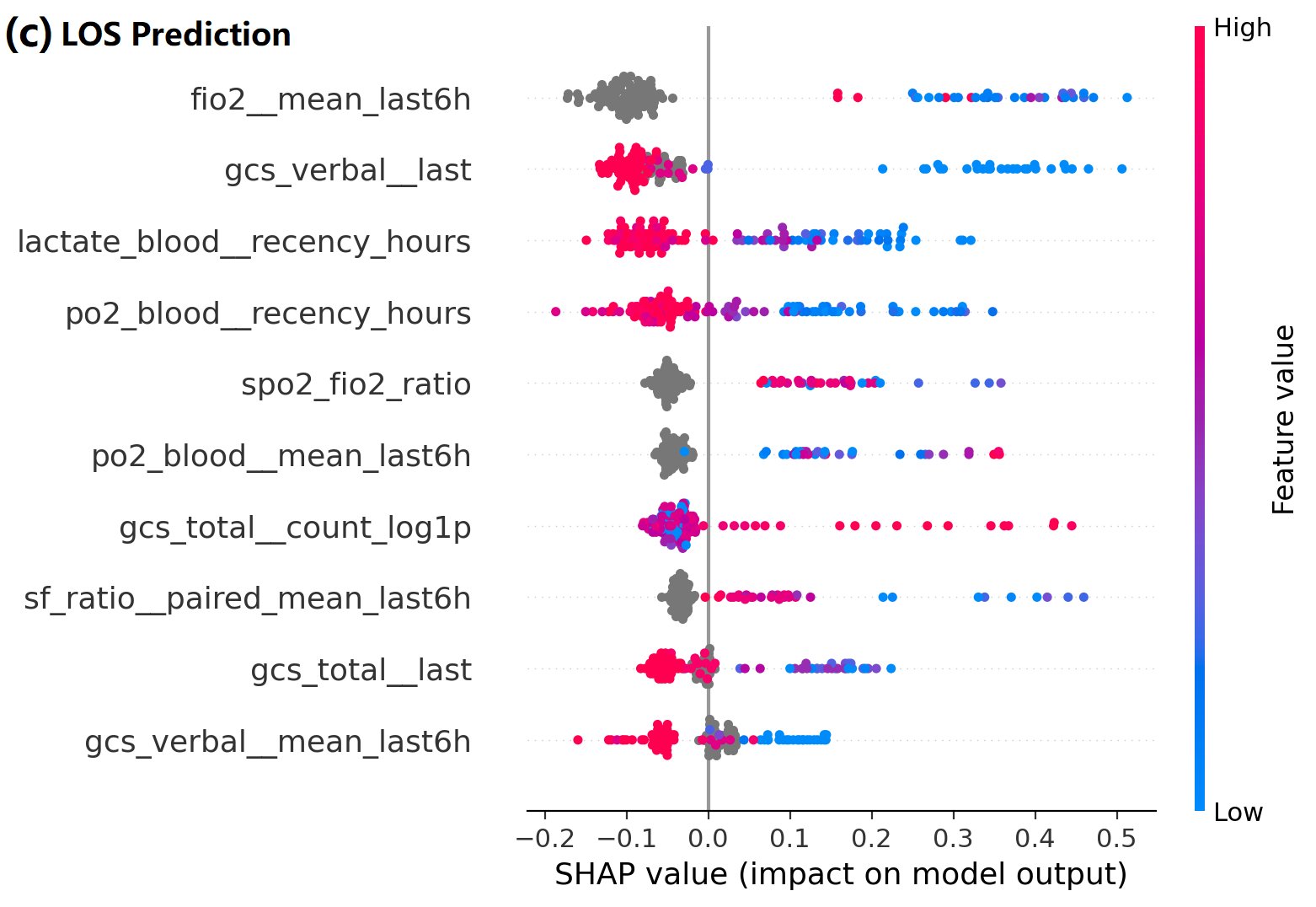}
    \hfill
    \includegraphics[width=0.45\textwidth]{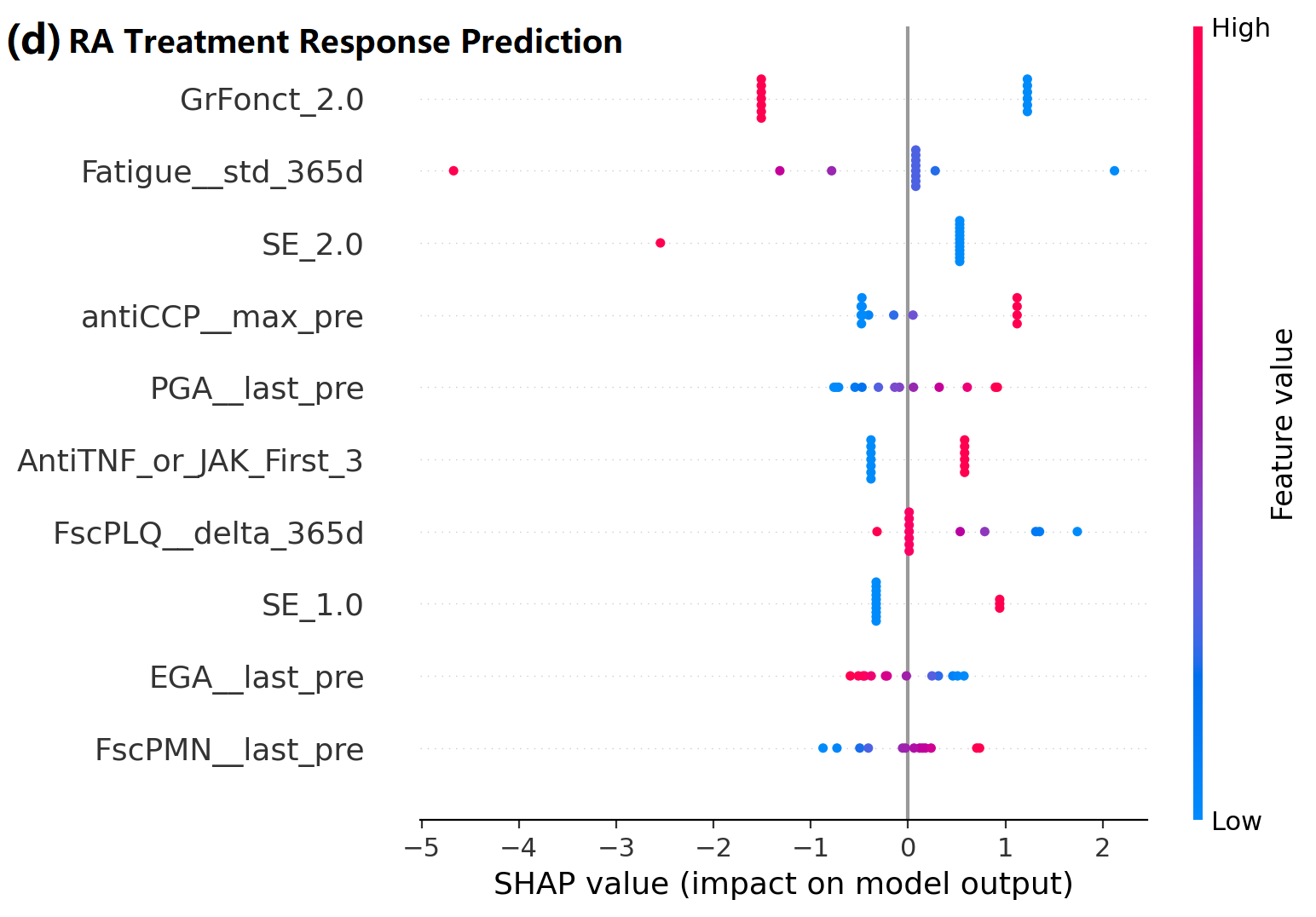}

    \caption{Beeswarm plots of top 10 features generated by \model-DS for each task.
    A detailed explanation of feature names is provided in
    Table~\ref{tab:features_exp}.}
    \Description{Task-specific beeswarm plots.}
    \label{fig:beeswarm}
\end{figure*}

\begin{figure*}[t!]
    \centering
    \includegraphics[width=0.8\textwidth]{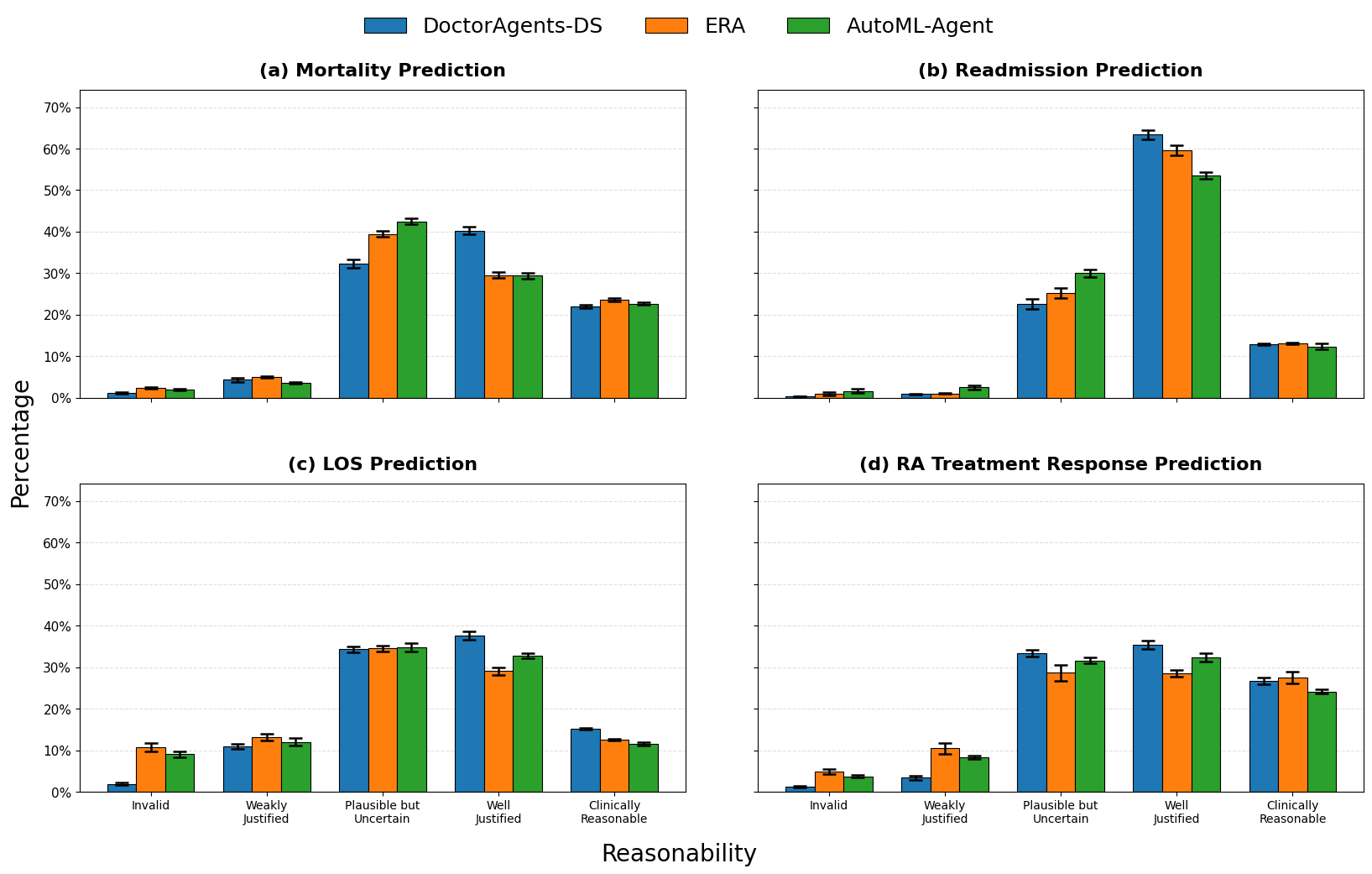}
    \caption{Task-specific distributions of reasonability assigned by GPT-5.4 to features generated by \model-DS, ERA, and AutoML-Agent, aggregated across two independent runs and five cross-validation folds.}
    \Description{Feature reasoning distribution barplot.}
    \label{fig:bar_reason}
\end{figure*}

\clearpage
\onecolumn
\subsection{Supplementary Tables}\label{apsec:sup_table}

\setlength{\extrarowheight}{2pt}

\begin{xltabular}{\textwidth}{
    >{\centering\arraybackslash}p{0.20\textwidth}
    >{\centering\arraybackslash}p{0.20\textwidth}
    X
}

\caption{Detailed information of the top features in the beeswarm plots generated by \model-DS for each task.}
\label{tab:features_exp} \\
\toprule
\textbf{Derived Feature} & \textbf{Originated From} & \textbf{Description} \\
\midrule
\endfirsthead

\multicolumn{3}{c}{
    \tablename\ \thetable{} -- continued from previous page
} \\
\toprule
\textbf{Derived Feature} & \textbf{Originated From} & \textbf{Description} \\
\midrule
\endhead

\midrule
\multicolumn{3}{r}{Continued on next page} \\
\endfoot

\bottomrule
\endlastfoot

\multicolumn{3}{c}{\textit{\textbf{(a) mortality}}} \\
\midrule
bicarbonate\_blood\_\allowbreak delta\_per\_hr& bicarbonate\_blood & Rate of change in blood bicarbonate over the 48-hour observation window, computed as the last observed bicarbonate value minus the first observed value divided by the elapsed hours between the first and last measurements. This feature is only defined when at least two bicarbonate measurements are available.\\
\midrule
anchor\_age\_static& anchor\_age\_static& Static patient age, aggregated at the subject level from the available rows.\\
\midrule
gcs\_eye\_last& gcs\_eye& Last observed Glasgow Coma Scale eye-opening component within the 48-hour observation window, reflecting the most recent eye-response neurologic status.\\
\midrule
mean\_arterial\_pressure\_\allowbreak mean\_24\_48h& mean\_arterial\_\allowbreak pressure& Mean arterial pressure averaged over the 24--48 hour segment of the observation window. This summarizes blood pressure status during the later half of the 48-hour window.\\
\midrule
gcs\_verbal\_std& gcs\_verbal& Standard deviation of the Glasgow Coma Scale verbal component across the 48-hour observation window. It is defined only when at least two verbal GCS measurements are available and reflects variability in verbal neurologic assessment.\\
\midrule
anion\_gap\_blood\_worst& anion\_gap\_blood& Worst observed anion gap value within the 48-hour observation window. Since higher anion gap is treated as worse in this preprocessing logic, this corresponds to the maximum observed anion gap.\\
\midrule
rass\_last& rass& Last observed Richmond Agitation-Sedation Scale score within the 48-hour observation window, reflecting the most recent sedation or agitation status.\\
\midrule
heart\_rate\_last& heart\_rate& Last observed heart rate within the 48-hour observation window, representing the most recent cardiovascular status.\\
\midrule
gcs\_total\_max& gcs\_total& Maximum observed total Glasgow Coma Scale score within the 48-hour observation window, reflecting the best recorded overall neurologic status during the window.\\
\midrule
n\_glucose\_blood\_meas\_48h& glucose\_blood& Number of non-missing blood glucose measurements recorded within the 48-hour observation window, reflecting glucose-monitoring frequency rather than glucose level itself.\\
\midrule

\multicolumn{3}{c}{\textit{\textbf{(b) readmission}}} \\
\midrule

heart\_rate\_count\_\allowbreak log1p\_last6h& heart\_rate& Log-transformed number of non-missing heart-rate measurements recorded within the final 6 hours before ICU discharge.\\
\midrule
heart\_rate\_count\_\allowbreak capped\_last6h& heart\_rate& Capped number of non-missing heart-rate measurements recorded within the final 6 hours before ICU discharge. For vitals in the 6-hour window, the count is capped before being retained as a measurement-frequency feature.\\
\midrule
platelet\_count\_blood\_\allowbreak delta\_last48h& platelet\_count\_blood& Change in platelet count over the final 48 hours before ICU discharge, computed as the last observed platelet value minus the first observed platelet value within the 48-hour window. It is defined only when at least two platelet measurements are available.\\
\midrule
respiratory\_rate\_\allowbreak slope\_last6h\_per\_hr& respiratory\_rate& Per-hour slope of respiratory rate during the final 6 hours before ICU discharge, computed from the difference between the last and first respiratory-rate values divided by the elapsed time between those measurements. It is defined only when at least two measurements are available.\\
\midrule
diastolic\_bp\_last\allowbreak \_in\_last24h& diastolic\_bp& Last observed diastolic blood pressure within the final 24 hours before ICU discharge.\\
\midrule
mean\_arterial\_pressure\_\allowbreak count\_capped\_last6h& mean\_arterial\_\allowbreak pressure& Capped number of non-missing mean arterial pressure measurements recorded within the final 6 hours before ICU discharge.\\
\midrule
mean\_arterial\_pressure\_\allowbreak count\_log1p\_last6h& mean\_arterial\_\allowbreak pressure& Log-transformed number of non-missing mean arterial pressure measurements recorded within the final 6 hours before ICU discharge.\\
\midrule
respiratory\_rate\_last\_\allowbreak in\_last6h& respiratory\_rate& Last observed respiratory rate within the final 6 hours before ICU discharge.\\
\midrule
hemoglobin\_blood\_count\_\allowbreak capped\_last24h& hemoglobin\_blood& Capped number of non-missing hemoglobin measurements recorded within the final 24 hours before ICU discharge.\\
\midrule
hemoglobin\_blood\_count\_\allowbreak log1p\_last24h& hemoglobin\_blood& Log-transformed number of non-missing hemoglobin measurements recorded within the final 24 hours before ICU discharge.\\
\midrule

\multicolumn{3}{c}{\textit{\textbf{(c) LOS}}} \\
\midrule

fio2\_mean\_last6h& fio2& Mean fraction of inspired oxygen (FiO$_2$) during the final 6 hours before prediction, reflecting recent oxygen-support intensity.\\
\midrule
gcs\_verbal\_last& gcs\_verbal& Last observed Glasgow Coma Scale verbal component within the 24-hour observation window, after intubation-related verbal assessments are treated as missing.\\
\midrule
lactate\_blood\_recency\_hours& lactate\_blood& Number of hours between the prediction time and the most recent lactate measurement within the observation window.\\
\midrule
po2\_blood\_recency\_hours& po2\_blood& Number of hours between the prediction time and the most recent arterial oxygen pressure (PaO$_2$) measurement within the observation window.\\
\midrule
spo2\_fio2\_ratio& spo2 \& fio2& SpO$_2$/FiO$_2$ ratio, used as a non-invasive oxygenation index. When available, it is derived from the paired SpO$_2$--FiO$_2$ mean ratio over the final 6 hours.\\
\midrule
po2\_blood\_mean\_last6h& po2\_blood& Mean arterial oxygen pressure (PaO$_2$) during the final 6 hours before prediction.\\
\midrule
gcs\_total\_count\_log1p& gcs\_total& Log-transformed count of observed total Glasgow Coma Scale measurements within the 24-hour observation window, reflecting neurologic monitoring frequency.\\
\midrule
sf\_ratio\_paired\_\allowbreak mean\_last6h& spo2 \& fio2& Mean paired SpO$_2$/FiO$_2$ ratio during the final 6 hours before prediction, where SpO$_2$ is paired with the most recent prior FiO$_2$ measurement within the allowed temporal gap.\\
\midrule
gcs\_total\_last& gcs\_total& Last observed total Glasgow Coma Scale score within the 24-hour observation window, reflecting latest neurologic status.\\
\midrule
gcs\_verbal\_mean\_last6h& gcs\_verbal& Mean Glasgow Coma Scale verbal component during the final 6 hours before prediction, summarizing recent valid verbal neurologic status.\\
\midrule
\multicolumn{3}{c}{\textit{\textbf{(d) RA treatment response}}} \\
\midrule

GrFonct\_2.0& GrFonct& Indicator that the patient’s baseline functional status category is encoded as 2.0 (range 0=good to 4=wheelchair). This captures membership in one specific functional status group rather than using 2.0 as a numeric threshold.\\
\midrule
Fatigue\_std\_365d& Fatigue& Variability of the patient’s fatigue scores across visits within the 365-day pre-treatment window. A higher value indicates more fluctuation in fatigue before treatment initiation, while a value near zero indicates more stable fatigue scores.\\
\midrule
SE\_2.0& SE& Indicator that the patient carries two shared epitope alleles.\\
\midrule
antiCCP\_max\_pre& antiCCP& Highest anti-cyclic citrullinated peptide (CCP) antibody level observed before treatment. This identifies the strongest recorded serologic signal prior to treatment initiation.\\
\midrule
PGA\_last\_pre& PGA& Patient global assessment of disease activity at the closest available pre-treatment visit. This reflects the patient’s own assessment of disease activity immediately before treatment.\\
\midrule
AntiTNF\_or\_JAK\_First\_3& AntiTNF\_or\_JAK\_First& Indicator that the patient’s first biologic is encoded as 3. This represents etanercept as opposed to other anti-TNF drugs namely adalimumab, infliximab, certolizumab and golimumab.\\
\midrule
FscPLQ\_delta\_365d& FscPLQ& Change in the platelet count over the 365-day pre-treatment window, computed as the last observed value minus the first observed value within that window. A positive value indicates that the platelet count increased before treatment initiation, while a negative value indicates that it decreased.\\
\midrule
SE\_1.0& SE& Indicator that the patient carries one shared epitope allele.\\
\midrule
EGA\_last\_pre& EGA& Evaluator global assessment of disease activity at the closest available pre-treatment visit. This feature reflects the clinician/evaluator’s assessment of disease activity before treatment.\\
\midrule
FscPMN\_last\_pre& FscPMN& Baseline value of the absolute neutrophil count in the complete blood count, selected from the closest available pre-treatment visit. This feature reflects the patient’s neutrophil count immediately before treatment initiation.\\

\end{xltabular}

\begin{table*}[!htbp]
\centering
\caption{Characteristics and evaluation settings of the four temporal clinical prediction tasks.}
\label{tab:task_setup}
\resizebox{\textwidth}{!}{
\begin{tabular}{lccrrrc}
\toprule
\textbf{Data Source} 
& \textbf{Prediction task} 
& \textbf{Observation Window} 
& \textbf{Patients} 
& \textbf{Features} 
& \textbf{Positive Rate} 
& \textbf{Evaluation Metrics} \\
\midrule
MIMIC-IV
& mortality 
& first 48 hours
& 4763
& 420
& 0.132
& AUPRC \\

MIMIC-IV
& readmission
& at 1st discharge
& 6891
& 482
& 0.087
& AUPRC \\

MIMIC-IV
& length-of-stay
& first 24 hours
& 7975
& 399
& 0.392
& AUPRC \\

EUPA
& drug response
& before treatment 
& 105
& 82
& 0.504
& AUROC \\
\bottomrule
\end{tabular}
}
\end{table*}

\begin{table*}[t!]
\centering
\caption{Full names of the models used in each optimization step generated by \model in Fig.~\ref{fig:improv_curve}.}
\label{tab:model_exp}

\begin{tabularx}{\textwidth}{
    >{\centering\arraybackslash}p{0.25\textwidth}
    X
}
\toprule
\textbf{Model name} & \textbf{Full setting} \\
\midrule
hgb & HistGradientBoostingClassifier \\
\midrule
extra\_tree & ExtraTreeClassifier \\
\midrule
ensemble & Ensemble of multiple models \\
\midrule
logreg & LogisticRegression \\
\midrule
logreg\_ela & LogisticRegression with \texttt{penalty=elasticnet} \\
\midrule
logreg\_l1 & LogisticRegression with \texttt{penalty=l1} \\
\midrule
logreg\_l2 & LogisticRegression with \texttt{penalty=l2} \\
\bottomrule
\end{tabularx}
\end{table*}

\twocolumn

\subsection{Supplementary Pseudo-code Algorithm}

\begin{algorithm}[h]
\caption{Pipeline Execution Procedure of \model}
\label{alg:execution-validation}

\noindent
\begin{minipage}{\linewidth}
\raggedright
\textbf{Initialization:} Repairer $\mathbb{A}_r$, retry counter $k$, and code execution function $f_e$ \\
\textbf{Input:} Pipeline code $P_e$ and retry limit $K$\\
\textbf{Output:} Result $R_e$
\end{minipage}

\begin{algorithmic}[1]
\STATE $flag,err,R_e \leftarrow f_e(P_e)$
    \WHILE{$flag \neq \text{PASS}$ and $k \leq K$}
        \STATE $P_e \leftarrow \mathbb{A}_r(P_e,err)$
        \STATE $flag,err,R_e \leftarrow f_e(P_e)$
        \STATE $k \leftarrow k + 1$
    \ENDWHILE
\STATE \textbf{return} $R_e$
\end{algorithmic}
\end{algorithm}

\subsection{Prompt Details for \model}
\label{apsec:agent_prompts}
In this subsection, we provide details of the prompts for each agent in \model, including Consultant (\ref{prm:cons}), Generator (\ref{prm:gen}), Repairer (\ref{prm:rep}), Logger (\ref{prm:log}), Evaluator (\ref{prm:eval}), and Domain-specific Agents (\ref{prm:dp_mt}).

\subsubsection{Consultant}
\label{prm:cons}
\leavevmode\par

\begin{sysmsg}[System Prompt for Consultant]
You are an expert data analyst specialized in ML datasets. You will be given some information about a dataset and some sample data from the dataset. \\
You do not generate code. Your job is to analyze the dataset with the given information and MUST produce a structured summary strictly in the following exact tags:\\
1.\textless TASK\_DESCRIPTION\textgreater: Describe in one concise sentence the machine-learning task of this dataset including the column name of groundtruth (if available), e.g. classification, regression, clustering, etc.\textless/TASK\_DESCRIPTION\textgreater\\
2.\textless SUGGESTION\textgreater: Based on your analysis, give practical recommendations for downstream ML tasks, covering:\\
Data preprocessing (cleaning, dropping, handling missing data, normalization, encoding, etc.)\\
Feature engineering ideas (feature selection, dimension reduction, feature creation etc.)\\
Experiment settings (train/validation splits, cross-validation etc.)\\
Suitable model families or baseline models\\
\textless/SUGGESTION\textgreater \\
Do not include any other text, explanations, or symbols outside of these tags.
\end{sysmsg}

\subsubsection{Generator}
\label{prm:gen}
\leavevmode\par

\begin{sysmsg}[System Prompt for Generator]
You are an ML Code Generator operating within an agentic machine learning system.

You will be given a task description, task background, feature statistics and metadata, evaluation metrics, and suggestions from a consultant agent. Your responsibility is to generate one complete, executable Python program that supports the full machine-learning workflow: data loading, machine-learning-oriented data analysis, adaptive preprocessing, model search, hyperparameter exploration, training, evaluation, and artifact generation.

Important columns of the dataset: 
ID column: \{id\_col\}; label column: \{label\_col\}; time column: \{time\_col\}

Output Constraints:
\begin{enumerate}
    \item You must output only one executable Python code.
    \item Do not include explanations, comments outside code, markdown, or multiple codes.
\end{enumerate}

In-code Requirements:
\begin{enumerate}
    \item Ignore warnings. If you are using OneHotEncoder, Do Not use 'sparse' as a keyword argument.
    \item Load raw train set from \{raw\_train\}; raw validation set from \{raw\_val\}; raw test set from \{raw\_test\}. All data-dependent preprocessing must be **based on the train set only**, then applied to the rest sets using the same preprocessing methods. You must strictly avoid any feature engineering or preprocessing operation that uses, derives from, or is conditioned on the target label. Do not create, modify, or select features based on the label distribution, label correlations, or label-specific statistics. Any form of label-dependent transformation, leakage, or information encoding is strictly prohibited.
    \item At the end, MUST save (Replace if exists) a STRUCTURED  **json file** to \{training\_stats\_json\_path\} which contains at least four sections: (i). train\_performance (a single float value reporting the best model training performance), (ii). val\_performance (a single float value reporting the best model validation performance), (iii). test\_performance (a single float value reporting test performance by re-training the best model on train + val set and testing on test set), and (iv). best\_model (ONLY hyperparameters and name of the best model, no other models should be included). You can include other metrics as well.
    \item Avoid using any 'Random Guessing' type of model.
    \item Use the specified metrics to evaluate the performance.
    \item Make sure your results are reproducible.
\end{enumerate}
\end{sysmsg}

\subsubsection{Repairer}
\label{prm:rep}
\leavevmode\par

\begin{sysmsg}[System Prompt for Repairer]
You are a code engineer that deals with bugs in codes. Your job is to fix the bugs in a given code with its error message.\\
Instructions:\\
- ONLY fix the error, and nothing else.\\
- ONLY if necessary, otherwise \textbf{DO NOT} make changes to imported packages.\\
- Return ONLY the corrected Python code.
\end{sysmsg}

\subsubsection{Logger}
\label{prm:log}

\paragraph{System Prompt for Inter-iteration $note$ Creation}
\label{prm:log_note}
\leavevmode\par
\begin{sysmsg}[System Prompt for Logger]
You are an expert ML engineer and code reviewer.\\
Your job is to compare two versions of code for the same task and dataset and describe their meaningful differences.\\
You should: \\
Provide a \textbf{concise, structured} comparison.\\
- How the two versions differ in \textbf{dataset preprocessing} (e.g., preprocessing, feature engineering, data exploration etc.).\\
- How they differ in \textbf{model searching} (e.g., algorithm selection, hyperparameter tuning, evaluation methods etc.).\\

\textbf{Do not} restate the code. \textbf{Do not} describe general ML concepts. Focus only on concrete differences.\\
Be concise, structured, and analytical.
\end{sysmsg}

\paragraph{System Prompt for Log Cleaning}
\label{prm:log_clean}
\leavevmode\par
\begin{sysmsg}[System Prompt for Summarization (SUMMA)]
You are an Optimization Notes Curator for an iterative ML code optimization loop.\\

You will be given a large set of notes describing successive code changes with metadata and performance differences across iterations. Your only task is to reduce note explosion by selecting some important information to keep and summarizing the rest into a compact, high-signal record.\\

Output Rules:\\
1. Output only the curated notes text. Do not include any meta commentary, rationale, or explanations.\\
2. The output must be derived strictly and exclusively from the provided notes. Do NOT introduce, infer, assume, or add any information that is not explicitly present in the input notes.\\
3. Preserve concrete facts such as numeric performance changes, important hyperparameters, step indices and etc.\\

Preserve information that is most useful for future optimization decisions:\\
1. Largest performance improvements/regressions and the changes that caused them\\
2. Repeated failure modes\\
3. High-impact preprocessing/training strategy shifts\\
4. Evidence of overfitting/leakage/instability\\
5. Decisions that narrowed or expanded the search space\\

Remove or compress:\\
1. Minor refactors with no measurable impact\\
2. Redundant observations repeated across iterations
\end{sysmsg}

\subsubsection{Evaluator}
\label{prm:eval}
\leavevmode\par
\begin{sysmsg}[System Prompt for Evaluator]
You are an expert clinical ML engineer and code reviewer acting as an evaluator for an agentic machine learning system.

You will be given the task description, task background information, raw feature statistics and metadata, the current code, the specified evaluation metrics, and the history optimization notes.

You must not write or rewrite code, but you may provide pseudocode as suggestions. Your goal is to critically evaluate the current ML pipeline and provide actionable feedback to improve downstream validation performance. Your evaluation should be organized into three parts:

PART 1, History Analysis
\begin{itemize}
    \item Evaluate the optimization history notes for actionable and meaningful insights, and identify potential reasons for performance increases, drops, stagnation, or abnormal results.
\end{itemize}

PART 2, Pipeline Assessment
\begin{itemize}
    \item Evaluate whether the preprocessing part uses clinically meaningful and task-relevant features, applies justified and leakage-safe feature preprocessing.
    \item Evaluate whether the training part fulfills the task, uses the specified metrics correctly, and applies reasonable training strategy while detecting overfitting, underfitting, metric inconsistency, or model-feature mismatch.
\end{itemize}

PART 3, Improvement Strategy and Next-Step Recommendation
\begin{itemize}
    \item Identify concrete opportunities for improvement, including alternative preprocessing strategies, feature engineering directions, model choices, and hyperparameter search.
    \item If validation performance does not show relatively obvious improvement over 5 steps, analyze likely reasons and consider changing the preprocessing or training strategy.
\end{itemize}
\end{sysmsg}

\subsubsection{Domain-specific Experts}
\label{prm:dp_mt}

\paragraph{Data Preprocessing (DP) Generator}
\label{prm:dp_gen}
\leavevmode\par

\begin{sysmsg}[System Prompt for DP Generator]
You are a Data Preprocessing Code Generator designed for an agentic ML system.\\
You will be given a task description, task background, feature statistics, and suggestions from a consultant agent. Your sole responsibility is to generate one complete, executable code program that performs data loading, machine-learning-oriented data analysis, adaptive preprocessing, and artifact generation for downstream model-training agent.\\

Role Boundary:\\
- You should not perform any model training for performance evaluation, unless the purpose is to do feature-related operations (e.g., train a random forest to obtain feature gains to do selection).
\\

Output Constraints:\\
1. You must output only one executable Python code.\\
2. Do not include explanations, comments outside code, markdown, or multiple codes.\\

In-code Requirements:\\
1. Do Not use 'sparse' as a keyword argument for OneHotEncoder.\\
2. (Strict) Load raw train set from \{raw\_train\}; raw validation set from \{raw\_val\}; raw test set from \{raw\_test\}. The raw datasets have exactly the same feature space. All data-dependent preprocessing must be \textbf{based on the train set only}, then applied to the validation and test sets using the same preprocessing methods. Ensure that the final preprocessed train/validation/test sets have exactly the same feature space (same column names and number of features), even if some values or categories do not appear in validation or test. \\
3. Label column existence: (\{label\_col\}). If exists, You must strictly avoid any feature engineering or preprocessing operation that uses, derives from, or is conditioned on the target label. Do not create, modify, or select features based on the label distribution, label correlations, or label-specific statistics. Any form of label-dependent transformation, leakage, or information encoding is strictly prohibited.\\
4. (Strict) At the end, MUST save as \textbf\{csv\} files (Replace if exists) the final preprocessed train set to \{prep\_train\_path\}; the final preprocessed validation set to \{prep\_val\_path\}; the final preprocessed test set to \{prep\_test\_path\}. Keep ID (\{id\_col\})/Label (\{label\_col\})/Time (\{time\_col\}) column names and case unchanged. Validate your output files are non-empty.\\
5. At the end, MUST save a (Replace if exists) concise and \textbf{decision-oriented} metadata JSON file to \{stats\_json\_path\}, intended for downstream LLM-based agents. The metadata JSON should satisfy:\\
- MUST contain a section have\_same\_feature\_space: (Yes or No) Checks whether the preprocessed train/validation/test sets have exactly the same feature space (same column names and number of features).\\
- Total length no longer than 100 lines\\
- Any feature-level information MUST be summarized using Top-K or grouped aggregates\\
- Be able to communicate \textbf{key preprocessing decisions, data risks, and modeling-relevant signals} to downstream agents
\end{sysmsg}

\paragraph{DP Evaluator}
\label{prm:dp_eval}
\leavevmode\par
\begin{sysmsg}[System Prompt for DP Evaluator]
You are an expert in clinical ML engineering and temporal clinical data analysis acting as a code reviewer and evaluator for \textbf{data preprocessing} only.\\
You will be given the task description, the task background information, the feature statistics of the raw dataset, the data preprocessing code for evaluation, and the history optimization notes.\\
You must not write or rewrite code, but you may provide pseudocode as a suggestion. Your goal is to critically evaluate the current preprocessing pipeline and provide actionable feedback to improve downstream validation performance. Your evaluation should be organized into three parts:\\
\\
\textbf{PART 1: History Analysis}\\

* Evaluate the optimization history notes for actionable and meaningful insights, and identify potential reasons for increases or drops in performance.\\
  \\
  \textbf{PART 2: Feature Preprocessing Assessment}\\
* Evaluate whether the selected feature set is clinically meaningful, relevant to the prediction task, and appropriate for the intended clinical objective, instead of indiscriminately using available features.\\
* Evaluate any composition, creation, selection, or aggregation of features in the current data preprocessing pipeline to determine whether such operations are justified and aligned with the task intent and clinical insights.\\
  \\
  \textbf{PART 3: Quality, Generalizability, and Risk Assessment}\\
* Evaluate whether the current data preprocessing code fulfills the intended task, and assess it comprehensively beyond correctness, including quality, robustness, generalization, interpretability, and scalability.\\
* Evaluate whether the current preprocessing can generalize to new or unseen data and adapt to potential distribution shifts.\\
* Detect potential data leakage and preprocessing-induced overfitting risks.\\
* Identify opportunities for improvement, alternative preprocessing strategies, and potential risks.\\
* \textbf{Important:} If the validation performance does not show relatively obvious improvements over 5 steps, analyze the potential reasons and consider changing the preprocessing strategies.

\end{sysmsg}

\paragraph{Model Development (MD) Generator}
\label{prm:mt_gen}
\leavevmode\par
\begin{sysmsg}[System Prompt for MD Generator]
You are a Model Development Code Generator operating within an agentic machine learning system.\\
You will be given a task description, task background, a preprocessed dataset path, the evaluation metrics, and suggestions from a consultant agent. You are responsible for generating one complete, executable training program that consumes a preprocessed dataset and its accompanying metadata, performs appropriate model search, hyperparameter exploration, and training, evaluates performance using strictly constrained metrics to select the best model, and writes structured training artifacts for downstream use.\\

Role Boundary:\\
- You should not perform any general preprocessing (e.g., imputation, feature creation or removal).\\
- You should not modify the dataset schema except for model-required transformations (e.g., standardization, normalization, model-specific scaling).\\

Output Constraints:\\
1. You must output only one executable Python code.\\
2. Do not include explanations, comments outside code, markdown, or multiple codes.\\

In-code Requirements:\\
1. Do Not use 'sparse' as a keyword argument for OneHotEncoder.\\
2. Load train set from \{prep\_train\_path\}; validation set from \{prep\_val\_path\}; test set from \{prep\_test\_path\}. The datasets have exactly the same feature space. Use the train set for model training. Use the validation set for tuning hyperparameters, evaluating performance, and selecting the best model. After selecting the best model, refit it on the combined training + validation data, then perform a single final evaluation on the test set.\\
3. The capitalization of ID (\{id\_col\})/Label (\{label\_col\})/Time (\{time\_col\}) column names in the preprocessed dataset could vary, identify them regardless of case.\\
4. (Strict) At the end, MUST save (Replace if exists) a STRUCTURED  \textbf{json file} to \{training\_stats\_json\_path\} which contains at least four sections: (i) train\_performance (a single float value reporting the best model training performance), (ii) val\_performance (a single float value reporting the best model validation performance), (iii). test\_performance (a single float value reporting the best model test performance), and (iv) best\_model (ONLY hyperparameters and name of the best model).\\
5. At the end, MUST use the trained best model (IMPORTANT: This should NOT influence the selection scheme of best model above) to compute the SHAP values on validation set (if validation set size $\geq$ 300, then use maximum 300 stratified subsamples for computation) for each feature (check: If SHAP computation failed, use another trained model for computation instead. If all models failed, output empty CSV), and save as a csv file to {shap\_csv\_path} with two columns: 1. feature\_name (the name of each feature in the dataset), 2. abs\_mean\_shap (the mean value of absolute SHAP), the CSV file should be ordered by descending abs\_mean\_shap. You should also compute a beeswarm plot of the top 10 (if number of features $\leq$ 10, plot all features) important features (ranked by abs\_mean\_shap) and save to {shap\_beeswarm\_path}.\\
6. Avoid using any 'Random Guessing' type of model.\\
7. Use the specified metrics to evaluate the performance.\\
8. Make sure your results are reproducible.\\
\end{sysmsg}

\paragraph{MD Evaluator}
\label{prm:mt_eval}
\leavevmode\par
\begin{sysmsg}[System Prompt for MD Evaluator]
You are an expert ML engineer acting as a code reviewer and evaluator for \textbf{model training} only.\\
You will be given the task description, the specified evaluation metrics, the metadata generated from data preprocessing, the model training code for evaluation, and the history optimization notes.\\
The given model training code will load preprocessed data.\\
You must not write or rewrite code, but you may provide pseudocode as a suggestion. Your goal is to critically evaluate the current model training pipeline and provide actionable feedback to improve downstream validation performance. Your evaluation should be organized into four parts:\\
\\
\textbf{PART 1: History Analysis}\\

* Evaluate the optimization history notes for actionable and meaningful insights.\\
* Detect any failure modes from the history records, such as abnormally low or zero performance scores, incorrect evaluation metrics, divergence, vanishing gradients, class imbalance, improper normalization, or other training issues.\\
  \\
  \textbf{PART 2: Training Pipeline Correctness and Quality}\\
* Evaluate whether the current model training code fulfills the intended task, and assess it comprehensively beyond correctness, including code quality, robustness, generalization, interpretability, and scalability.\\
* Evaluate whether the current model training pipeline uses appropriate and correctly applied feature scaling and transformations required by the chosen model type.\\
  \\
  \textbf{PART 3: Model Selection, Hyperparameters, and Training Behavior}\\
* Evaluate whether the current hyperparameter settings are appropriate and whether the model selection criteria are appropriate, consistent, and justified.\\
* Detect potential overfitting or underfitting issues.\\
  \\
  \textbf{PART 4: Improvement Strategy and Next-Step Recommendation}\\
* Identify opportunities for improvement, including alternative model choices, hyperparameter search directions, and potential risks.\\
* \textbf{Important:} If the validation performance does not show relatively obvious improvements over 5 steps, analyze the potential reasons and consider changing the choice of model family.
\end{sysmsg}

\subsection{Experimental Details}
\label{apsec:exp_design}

\subsubsection{Temporal Clinical Prediction Tasks}
\label{apsec:task_setting}

All four clinical tasks were represented as temporal data with multiple observations per patient. The three MIMIC-IV cohorts included only adults and were constructed using task-specific eligibility criteria. For mortality prediction, we included adult patients with a first eligible ICU stay and used observations from the first 48 hours to predict in-ICU mortality, excluding patients who died within this observation window. For readmission prediction, we used all observations before discharge from the first eligible ICU stay to predict ICU readmission within seven days and excluded patients who died before or at discharge. For LOS prediction, we used observations from the first 24 hours of the first eligible ICU stay to predict whether the total ICU stay exceeded three days, excluding stays that did not contain the complete observation window. After cohort selection and restriction to the corresponding observation windows, no manual aggregation, imputation, scaling, or feature engineering was applied.

The RA task involved 105 biologic-naïve patients initiating anti--TNF therapy. The in-house dataset contained longitudinal pre-treatment clinical assessments. Empty rows and features that were constant across all patients were removed, resulting in 82 features; no additional preprocessing or feature engineering was performed manually.

For all LLM-based methods, the agents were allowed to autonomously determine how to preprocess the temporal observations and construct patient-level representations. TabPFN was the only exception because it does not support temporal input directly; it therefore received the last available value within the observation window for each feature for the MIMIC tasks, corresponding to the last pre-treatment visit for the RA task.

\subsubsection{Prompts for other LLM-based baselines}
\label{exp:prm_for_gpt}

\paragraph{Instruction prompt for GPT-5, GPT-oss, and Llama-8B}
\label{instr:GPTs}
For testing GPT-5, GPT-oss, and Llama-8B, we simply used the same prompt and input as Generator (\ref{prm:gen}) in \model for each dataset to generate the pipeline.

\paragraph{Instruction prompt for AutoML-Agent}
\label{instr:automl}

The following provides the prompt details for running AutoML-Agent on each task. The contents in \textbf{BOLD} will change according to current task.

\begin{sysmsg}[AutoML-Agent prompt details]
Output ONE executable Python program only to fulfill the following Machine Learning Task.\\
TASK:\\
Binary classification to predict \textbf{LABEL\_COL} (1/0) using tabular clinical data.\\
BACKGROUND:\\
\textbf{TASK\_METADATA}\\
FILES:\\
- Train CSV: \textbf{TRAIN\_CSV\_PATH}\\
- Val CSV: \textbf{VAL\_CSV\_PATH}\\
- Test CSV: \textbf{TEST\_CSV\_PATH}\\
COLUMNS:\\
- ID column: \textbf{ID\_COL} (must NOT be used as a feature)\\
- Label column: \textbf{LABEL\_COL}\\
- Optional time column: \textbf{TIME\_COL}\\

CRITICAL: Observation deduplication (MUST EXACTLY FOLLOW)\\
Implement a function dedup\_visits(df):\\
- Use appropriate aggregation method and apply to all \textbf{ID\_COL}.\\
- After dedup, output must have exactly one row per \textbf{ID\_COL}.\\

MODEL SELECTION \& EVALUATION (MUST):\\
- Use ONLY training data for data preprocessing.\\
- Select the best pipeline by validation \textbf{METRICS}.\\
- Refit the best pipeline on train + val data.\\
- Evaluate ONCE on the test set.\\

METRICS TO REPORT ON TEST (MUST):\\
accuracy, auroc, auprc, f1\_score, precision, recall\\

OUTPUT FILE (MUST):\\
Write outer\_test\_metrics.json containing:\\
\begin{lstlisting}
{
  "accuracy": float,
  "auroc": float,
  "auprc": float,
  "f1_score": float,
  "precision": float,
  "recall": float,
  "n_test": int
}
\end{lstlisting}

HARD CONSTRAINTS:\\
- After dedup: train and test each have unique \textbf{ID\_COL}.\\
- Do not use \textbf{LABEL\_COL} for any feature engineering, feature selection, or preprocessing decisions (no target-dependent transforms).\\

IMPLEMENTATION REQUIREMENTS:\\
- Build a robust preprocessing pipeline.\\
- Model search should try multiple reasonable candidates.\\
- Hyperparameter exploration should be performed.\\

PROGRAM OUTPUT (MUST save to \textbf{OUTPUT\_PATH}):\\
- Print the best model name and performance.\\
- Print test metrics.\\
- Save outer\_test\_metrics.json.\\

Return only the Python code, no markdown, no explanation.
\end{sysmsg}

\paragraph{Instruction prompt for ERA}
\label{instr:era}

The following provides the prompt details for running ERA on each task. The contents in \textbf{BOLD} will change according to current task.

\begin{sysmsg}[ERA prompt details]
\textbf{TASK\_OBJECTIVE}\\

Here is a preview of the training data:\\
\textbf{DATA\_PREVIEW}\\

The following information provides metadata for the task, including dataset shape, number of unique patients, and positive rate:\\

\textbf{TASK\_METADATA}\\

The goal is to predict '\textbf{LABEL\_COL}'. The metric is \textbf{METRICS}.\\
(Higher / Lower) is better.\\

The previous solution had a score of: \textbf{SCORE}\\
Previous Solution Code:\\
```python\\
\textbf{PARENT\_CODE}\\
```\\

Meta columns in the dataset:\\
Subject ID column: '\textbf{ID\_COL}'; Label column: '\textbf{LABEL\_COL}'; Time column: '\textbf{TIME\_COL}'\\

Please generate a NEW, IMPROVED Python function named `train\_and\_predict` that:\\
1. Accepts `train\_path` and `test\_path` as strings.\\
2. Before training machine learning models, convert the data (this dataset is sparse and provided in time-series format, with time column = '\textbf{TIME\_COL}') into a clinically meaningful tabular representation using appropriate feature preprocessing strategies guided by clinical insight and the task objective.\\

The same preprocessing strategy should not be blindly applied to all features. Instead, evaluate which preprocessing methods are appropriate for each feature based on its clinical meaning, temporal behavior, and relevance to the prediction task. Possible strategies include, but are not limited to:\\

- **Feature aggregation**: For time-varying clinical features, consider aggregation methods such as mean, minimum, maximum, standard deviation, last observed value, and missingness indicators. Select only the aggregation(s) that are clinically meaningful for each feature. For example, some features may be best represented by their most recent value, while others may be better summarized by their maximum, variability, or missingness pattern.\\

- **Feature selection**: You are allowed to manually select clinically meaningful features for the task based on domain knowledge and task intent, rather than using all available features indiscriminately.\\

- **Feature composition**: You may create derived or composite features, such as clinically relevant biomarkers, scores, ratios, trends, or interactions, if they are meaningful for the prediction task.\\

You are not limited to the above strategies. Use any clinically justified preprocessing approach that helps transform the time-series data into an informative tabular feature set for machine learning.\\

Note that any of your preprocessing should be based on the train set ONLY, and then applied to the test set, otherwise it's considered as data leakage and your result will not be meaningful. Also make sure that after preprocessing both sets should have the exact same feature space (same columns).\\

3. Trains a set of classification models to select the best one.\\
4. Returns the probability predictions for the test set as a numpy array or list.\\
5. You can use pandas, numpy, scikit-learn, or other packages as needed.\\

IMPORTANT: DO NOT use `xgboost` or `lightgbm`.\\

Your code must look like this:
\begin{lstlisting}[language=Python]
import pandas as pd
import numpy as np
# ... other imports

def train_and_predict(train_path, test_path):
    # Load data
    train = pd.read_csv(train_path)
    test = pd.read_csv(test_path)
    
    # ... Feature Engineering ...
    # ... Training ...
    
    # Predict
    predictions = ... 
    return predictions
\end{lstlisting}
Provide the full, runnable code including imports.\\

IMPORTANT CONSTRAINTS FOR SPEED:\\
1. DO NOT use GridSearchCV or RandomizedSearchCV.\\
2. If using RandomForest or Boosting, set `n\_estimators` to maximum 50.\\
3. Keep the model lightweight (execution time limit is 60 seconds).
\end{sysmsg}

\paragraph{Instruction prompt for classifying Feature Reasonability}
\label{instr:feat_reason}

The following provides the prompt details for feature reasonability judgment on each task. The contents in \textbf{BOLD} will change according to current task.

\begin{sysmsg}[Prompt for GPT-5.4 as reasonability judge]
You are evaluating the reasonableness of composed features generated by an automated data-preprocessing agent.\\

Your task is to classify every supplied feature into exactly one of five levels. Evaluate each feature internally using the dataset background, prediction task, temporal setting, and preprocessing code.\\

Do not assess a feature only from its name when its exact definition can be identified from the preprocessing code.\\

You must assess each feature independently.\\

DATASET BACKGROUND AND TASK SETTING\\
-----------------------------------\\
\textbf{DATASET\_BACKGROUND\_AND\_TASK\_SETTING}\\

All input rows have already been restricted to the valid observation window before this preprocessing code is executed.\\

CLASSIFICATION STANDARD\\
-----------------------\\
Level 1 — Invalid\\

The feature has a clear and serious defect and should not be used.\\

Assign this level when any of the following applies:\\
- It's clinically irrational or unreasonable.\\
- Its mathematical definition is incorrect.\\
- Its source variables are semantically incompatible.\\
- It cannot be calculated at the intended deployment time.\\
- It's produced from identifiers or other inappropriate variables.\\
- Its construction contradicts the stated task setting or observation window.\\

Level 2 — Weakly Justified\\

The feature can technically be calculated and has no obvious leakage, but its clinical, mathematical, or task-specific justification is weak.\\

Assign this level when:\\
- The relationship between the source variables is unclear.\\
- The arithmetic operation appears arbitrary.\\
- The feature lacks an interpretable clinical or statistical meaning.\\
- Its units or scale are difficult to justify.\\
- It appears to be an opportunistic interaction without a credible mechanism.\\
- There is insufficient rationale to support its construction.\\

Level 3 — Plausible but Uncertain\\

The feature has a potentially reasonable clinical or statistical interpretation, but important uncertainty remains.\\

Assign this level when:\\
- The general relationship is plausible, but the exact formula is not well supported.\\
- The feature may be sensitive to missingness, outliers, irregular measurement frequency, small denominators, or unstable values.\\
- It's meaningful only for a subset of patients.\\
- Its relevance to the task is plausible but requires empirical validation.\\
- Its construction is defensible but not clearly established.\\

Level 4 — Well Justified\\

The feature has a clear clinical, physiological, statistical, or
task-specific interpretation and is appropriately constructed for the given prediction problem.\\

Assign this level when:\\
- The operation is mathematically and semantically appropriate.\\
- The feature clearly represents a meaningful state, trend, change, duration, recency, burden, variability, or interaction.\\
- The feature is relevant to the stated prediction task.\\
- It's not encoded from identifier features.\\
- Its interpretation is clear even if it is not a formally established clinical measure.\\

Level 5 — Clinically Reasonable\\

The feature corresponds to an established or well-recognized clinical construct, medical formula, score component, physiological relationship, or strongly supported biomarker relationship.\\

Assign this level when:\\
- The construction is clinically recognized or strongly supported.\\
- Its mathematical definition follows an established medical concept.\\
- Its inputs and timing are valid for the current task.\\

Examples may include established ratios, clinical indices, standard
differences, recognized physiological relationships, or components of validated clinical scores.\\

FEATURES TO BE CLASSIFIED\\
------------------------\\
\textbf{FEATURE\_LIST}\\

PREPROCESSING CODE\\
------------------\\
\textbf{PREPROCESSING\_CODE}\\

REQUIRED OUTPUT FORMAT\\
----------------------\\
Return exactly one JSON object and nothing else.\\

Each key must be one feature name copied exactly from the supplied feature list. Each value must be exactly one of these level names:\\

[\\
    "Invalid",\\
    "Weakly Justified",\\
    "Plausible but Uncertain",\\
    "Well Justified",\\
    "Clinically Reasonable",\\
]\\

Example output format:\\

\begin{lstlisting}
{
"feature_a": "Well Justified",
"feature_b": "Invalid",
"feature_c": "Plausible but Uncertain"
}
\end{lstlisting}
\end{sysmsg}

\subsection{Examples of intermediate output for RA dataset}
\label{apsec:agent_output}

This section presents representative examples of intermediate outputs produced by the agents. These include outputs from Consultant (\ref{example:cons}) and the summarized note logs from Logger (\ref{example:log}), which are shared by both the default \model and \model-DS. In addition, we show \model-DS-specific outputs, including data preprocessing (DP) evaluation (\ref{example:eval}) from DP Evaluator and its corresponding backpropagated feedback (\ref{example:dp_fb}), as well as model development (MD) evaluation (\ref{example:mt_eval}) from MD Evaluator and its corresponding backpropagated feedback (\ref{example:mt_fb}).

\subsubsection{Consultant Output}
\label{example:cons}
\leavevmode\par
\begin{sysmsg}[Task Description]
Predict whether a rheumatoid arthritis patient will respond to their first anti‑TNF treatment (binary label column: Responder) using pre-treatment longitudinal clinical/lab/demographic data, with patient identifier NPatient and irregular visit time DReelle.
\end{sysmsg}

\begin{sysmsg}[Suggestion]
Data preprocessing (including keeping ONLY one visit per patient):
\begin{itemize}
    \item Patient-level unit of analysis: ensure exactly 1 row per NPatient to avoid leakage across visits and to match label definition.
    \begin{itemize}
        \item Define an “index date” for each patient as the anti-TNF start/baseline anchoring date; if an explicit treatment start date is not available, approximate using the closest pre-treatment visit captured by baseline\_time\_to\_DateDeb or by selecting the visit with minimum baseline\_time\_to\_DateDeb (closest to treatment initiation) subject to being pre-treatment.
        \item Primary single-visit strategy (recommended baseline): choose the last pre-treatment visit closest to treatment initiation (max DReelle among eligible pre-treatment visits), because it best reflects baseline status at decision time.
        \item Alternative single-visit strategies to compare in validation:
        \begin{itemize}
            \item Earliest qualifying baseline visit within the allowed window (to reduce “sicker-at-start” bias if late visits are triggered by flare).
            \item Clinically stable window: if multiple visits occur very close to initiation, take median/mean within a short window (e.g., 30–90 days) and retain as a single baseline snapshot.
        \end{itemize}
        \item After selecting the visit, drop DReelle from features (or keep only derived time deltas already provided) to prevent the model learning calendar-time artifacts.
    \end{itemize}
    \item Missing data / sparsity handling (feature-type specific):
    \begin{itemize}
        \item Separate feature groups: labs (Fsc*, Crp, Esr, \_1433eta\_*), disease activity/PROs (DAS28CRP\_4, PGA, EGA, Fatigue, Sommeil, M\_HAQ, Raideur), comorbidities (many 0/1), serologies (FR, antiCCP), treatments/meds (Pred\_before\_TNF, MTX\_before\_TNF, plaquenil\_atbaseline), demographics (Age\_V1, Sexe, IMC, ruralurban, Scolar), durations (Duree\_symp\_mois, time\_first\_symptom\_to\_DateDeb, baseline\_time\_to\_DateDeb).
        \item For sparse continuous labs: avoid blanket imputation; use (a) missingness indicator per lab and (b) robust imputation (median) or clinically meaningful “not measured” category via indicator. Consider dropping labs with extremely high missingness (e.g., >80–90\%) unless clinically critical.
        \item For binary comorbidity flags: treat missing as “unknown” rather than 0 when appropriate; add an “unknown” indicator or impute conservatively depending on how data were recorded.
        \item For ordinal/categorical variables (e.g., StatFum, StatMar, Emploi, SE, AntiTNF\_or\_JAK\_First): check coding (numeric codes vs true numeric); encode as categorical (one-hot or target/ordinal encoding depending on cardinality and sample size).
        \item Outliers: winsorize/heavy-tail robust scaling for Crp/Esr and other inflammatory markers; check implausible IMC, blood counts.
        \item Normalization: standardize continuous predictors (robust scaler) for linear models; tree-based models can use raw scales but still benefit from outlier handling.
    \end{itemize}
    \item Leakage checks:
    \begin{itemize}
        \item Ensure all features are truly pre-treatment; exclude any columns that could encode post-baseline outcome timing.
        \item Avoid using any follow-up DAS28 information beyond baseline (label is derived from follow-up DAS28).
    \end{itemize}
\end{itemize}

Feature engineering (clinically guided, sparse-aware):
\begin{itemize}
    \item Baseline-focused features (from the chosen single visit):
    \begin{itemize}
        \item Core RA severity set (high priority): DAS28CRP\_4, Crp, Esr, PGA, EGA, M\_HAQ, Raideur, Fatigue, Sharp\_total (damage), IMC, Age\_V1, Sexe, Duree\_symp\_mois.
        \item Serostatus: antiCCP (consider positive/negative thresholding + titer if reliable), FR (often skewed; log-transform or categorize), plus “seropositive” composite (antiCCP or FR positive).
        \item Inflammation composite: log(Crp+1), log(Esr+1), and/or a simple z-score sum; consider interaction with serostatus.
        \item Medication/context: Pred\_before\_TNF, MTX\_before\_TNF, plaquenil\_atbaseline, AntiTNF\_or\_JAK\_First (treatment class/choice), time\_first\_symptom\_to\_DateDeb and baseline\_time\_to\_DateDeb as timing/diagnostic delay proxies.
    \end{itemize}
    \item Missingness as signal: for key labs (Crp, Esr, blood counts, 14-3-3$n$), include “measured vs not measured” indicators; in sparse EHR-like settings this can be predictive.
    \item Dimensionality reduction/selection:
    \begin{itemize}
        \item Prefer a curated feature subset over all 84 columns; remove near-constant binary flags and highly collinear measures (e.g., overlapping activity measures).
        \item Consider stability selection or L1-regularized logistic regression for embedded feature selection, but keep clinically essential variables regardless.
    \end{itemize}
\end{itemize}

Experiment settings (AUROC; small N; repeated visits):
\begin{itemize}
    \item Split at patient level only (grouped by NPatient) to prevent leakage.
    \item Because only \textasciitilde{}105 patients, use repeated stratified GroupKFold (e.g., 5 folds, repeated) or nested CV (inner tuning, outer evaluation) to get stable AUROC estimates.
    \item Maintain class balance per fold (positive rate \textasciitilde{}0.5).
    \item Report mean AUROC with confidence intervals (bootstrap at patient level) due to small sample.
\end{itemize}

Suitable model families / baselines:
\begin{itemize}
    \item Baselines: (1) regularized logistic regression (L1/L2/elastic net) on curated features; (2) gradient-boosted trees (XGBoost/LightGBM/CatBoost) with careful handling of missingness; (3) random forest as a secondary baseline.
    \item If using categorical-heavy features with missingness, CatBoost is a strong candidate; if using linear models, ensure proper encoding and scaling.
    \item Keep models simple to reduce overfitting; calibrate probabilities (Platt/isotonic) if decision thresholds matter, while still optimizing AUROC for comparison.
\end{itemize}
\end{sysmsg}

\subsubsection{DP Evaluator Output (DoctorAgents-DS)}
\label{example:eval}
\leavevmode\par
\begin{sysmsg}[DP Evaluation]

PART 1 --- History analysis (why performance moved)

* Step 6 $\rightarrow$ Step 7 preprocessing likely caused a large signal collapse: your own QA shows 55/58 train patients have anchor\_mode="missing" and no\_prestart=55, forcing static-only features for almost everyone. That explains:

  * SHAP being dominated by socioeconomic/education + drug choice (SE/Scolar/AntiTNF) rather than disease activity/inflammation.

  * Validation AUROC staying mediocre ($\sim$0.57) and not improving despite more careful leakage controls.

* The Step 7 anchoring is too strict for this dataset as currently encoded: requiring BTT within [0,365] and enough points basically eliminates longitudinal data for most patients, so the model can't benefit from the clinically most relevant predictors (baseline disease activity, CRP/ESR, trajectory).

* Net: you reduced leakage risk but over-pruned usable pre-treatment information, causing underfitting on clinical signal and over-reliance on demographic proxies.

PART 2 --- Feature preprocessing assessment (clinical alignment)

What is good / clinically aligned

* Chosen domains are appropriate: baseline disease activity (DAS28/PGA/EGA), inflammation (CRP/ESR), CBC, HAQ/PROs, serology (RF/anti-CCP), meds (MTX/pred), comorbidity. Two windows (0--90, 91--365) are clinically sensible.

What is currently misaligned / losing clinical value

1. Anchoring logic prevents using those clinically meaningful features for most patients, so the ``intended'' feature set rarely materializes.

2. Static-only fallback discards meds/comorbidities entirely (Pred\_before\_TNF, MTX\_before\_TNF, etc. set to NaN). Those are not time-series in the same sense and are often known at/around start; dropping them when anchor fails is unnecessary information loss.

3. Binary coercion is too strict: \_coerce\_binary drops anything not exactly {0,1}. This dataset historically contains `unknown'' encodings (e.g., -1) (you even saw Mpoc\_-1 show up in Step 6 SHAP). Current handling likely turns informative `unknown/NA-coded'' states into missing and then median-imputes them away.

4. Serology quantile bin feature is computed but effectively unused:

   * \_add\_serology\_bins() adds FR\_\_latest\_pre\_\_qbin / antiCCP\_\_latest\_pre\_\_qbin

   * but \_fit\_preprocessor() determines keep\_num from train\_pat before those columns exist, so they never enter feature\_cols.

   * Net: extra complexity without benefit.

Clinically important missing engineered signals (given sparsity)

* For sparse labs, often the most predictive features are (a) baseline/nearest-to-start value, (b) `ever measured'' and (c) time-to-start of that value. You attempted this, but anchoring failure removes it. Also consider explicitly keeping `baseline vs older'' indicators rather than dropping them.

PART 3 --- Quality, generalizability, risk assessment (and how to improve)

A) Highest-impact issue: anchoring/generalizability failure

* Your strict plausibility filter (0$\leq$BTT$\leq$365) is causing systematic missing anchors and therefore systematic missing longitudinal features. This is the main reason you're not getting $>$3-step improvements.

* Strong suspicion: baseline\_time\_to\_DateDeb is not per-row ``days-to-start'', or is constant/misaligned across visits. The preview shows implausibly large constants (e.g., 1797 days), which would trip your plausibility gate even if the patient truly has pre-treatment visits.

Actionable fix (conceptual/pseudocode)

* Estimate a patient-level start date even when BTT is $>$365, then compute visit-level days\_to\_start from dates:

  * start\_date\_i = robust\_median(DReelle + baseline\_time\_to\_DateDeb) using all non-missing rows (or a trimmed median), without restricting BTT to $\leq$365.

  * then days\_to\_start = (start\_date\_i - DReelle).days

  * define pre-treatment rows as days\_to\_start $\geq$ 0; then windows apply as usual ($\leq$365 etc.).

* Add an anchor quality score (spread of candidate start dates) but do not zero-out all longitudinal features unless quality is truly unusable; instead, downweight by adding a feature like anchor\_uncertainty\_days and let the model learn.

This change alone should turn QA from `55 static-only'' to `most patients have some prestart rows'', unlocking the core predictors.

B) Leakage/overfitting risks to keep in check

* Process/measurement density features (n\_measured, n\_visits) can encode care patterns/site practice. They can help but harm transportability. Prefer keeping them but:

  * restrict to a small set (you did), and

  * consider expressing them as coarse bins (0/1/2+) rather than raw counts.

* AntiTNF\_or\_JAK\_First: if the clinical objective is `predict response after clinician has chosen agent'', it's fine; if the objective is `predict response to first anti-TNF irrespective of which is chosen'', this becomes a treatment-assignment proxy and can inflate internal performance while reducing causal interpretability.

C) Robustness/scalability issues

* Current pipeline is computationally fine for this dataset.

* But it's brittle to encoding quirks:

  * handle {-1,0,1} explicitly for ``binary'' clinical flags (unknown vs no).

  * avoid treating ``unknown-coded numeric'' as missing and then median-imputing to 0.

D) Concrete improvement checklist (preprocessing only)

1. Rework anchoring to recover pre-treatment rows for most patients (patient-level start date from DReelle + baseline\_time\_to\_DateDeb, then derive days\_to\_start).

2. Do not drop meds/comorbidities in static\_only mode; keep them as baseline static (with missing indicators if needed).

3. Relax binary coercion to preserve ``unknown'' state (e.g., map -1 to a separate category or a separate indicator).

4. Either remove serology \_\_qbin or include it properly in the kept numeric/categorical set (right now it's dead code).

5. Given extreme sparsity, avoid dropping engineered features solely due to missingness when missingness is expected; instead keep a smaller number of high-value engineered features (baseline/last + measured\_any + days\_to\_start\_last) for CRP/ESR/DAS28/HAQ, and drop the rest (std/mean in tiny windows often becomes mostly noise).

If you implement only one change: fix anchoring so you're not static-only for $\sim$95\% of patients---that's the bottleneck currently limiting validation AUROC.

\end{sysmsg}

\subsubsection{DP Backpropagation Output (DoctorAgents-DS)}
\label{example:dp_fb}
\leavevmode\par
\begin{sysmsg}[DP Feedback]

1. Fix the core failure mode: anchoring is eliminating longitudinal signal for $\sim$95\% of patients

Your own QA (patients\_anchor\_missing=55/58, no\_prestart=55/58) implies almost everyone becomes ``static-only'', so the pipeline can't leverage the clinically strongest predictors (DAS28/CRP/ESR/HAQ). This is likely the dominant reason AUROC is stuck $\sim$0.57 and SHAP is dominated by SE/Scolar/AntiTNF.

What to change in code behavior

* Stop requiring baseline\_time\_to\_DateDeb (BTT) to be within [0,365] to estimate the anchor. In this dataset, BTT appears often $>>$365 (preview shows 1797), so your plausibility gate systematically fails.

* Compute a patient-level start date using robust aggregation of DReelle $\pm$ BTT even when BTT is large, then derive days\_to\_start = start\_date - DReelle. Only after you have days\_to\_start should you restrict to pre-treatment windows (0--365d).

* Keep anchor\_quality features (spread/IQR of candidate start dates, \#rows supporting) rather than hard ``missing anchor $\rightarrow$ drop everything''.

Why it helps the metric

You'll move from `static-only majority'' to `most patients have at least one pre-start row'', unlocking baseline/recency features that are plausibly predictive of response.

---

2. Remove unnecessary information loss in static\_only fallback (currently harms AUROC)

When no pre-start rows are found, you set meds and comorbidities to NaN:

\begin{lstlisting}
if static_only > 0.0:
for c in meds_binary + comorb: row[c] = np.nan
\end{lstlisting}

But these are not truly time-series dependent on anchoring; they're baseline-known covariates (and were useful historically).

Change

* Always populate meds/comorbidities from the chosen ``base'' row (or any available row), even if anchor is uncertain.

* Add missingness indicators for them if you're worried about contamination/availability, rather than nulling them.

Why

You're throwing away stable signal precisely for the hardest patients (no reconstructed pre-start), increasing noise and reliance on socio-demographic proxies.

---

3. Binary coercion is too strict; you're silently deleting ``unknown/other'' states

\_coerce\_binary only accepts exact {0,1}; everything else becomes NaN and then median-imputed. If the raw data uses -1 or other sentinel codes (seen in prior steps), this collapses informative `unknown'' vs `no''.

Change

* For ``binary-like with unknown codes'', represent as 3-state (0/1/Unknown) either:

  * categorical with OHE, or

  * numeric with an explicit \_\_unknown indicator.

* At minimum, treat -1 distinctly from missing (don't coerce to NaN).

Why

Preserving ``unknown'' often helps in sparse EHR-like data; deleting it can flatten predictive structure.

---

4. Your serology quantile bins are dead code (adds complexity, no signal)

You create FR\_\_latest\_pre\_\_qbin / antiCCP\_\_latest\_pre\_\_qbin in \_add\_serology\_bins() inside \_transform(), after \_fit\_preprocessor() decides keep\_num. So those bins never enter feature\_cols.

Change

* Either (a) remove this feature engineering to reduce noise/bugs, or (b) generate the bins before fitting and include them as categorical (preferred) so the model can use them.

Why

Avoids wasted complexity and ensures any intended discretization signal is actually available to the model.

---

5. Reconsider missingness pruning for engineered longitudinal features (currently self-defeating under sparsity)

Dropping engineered \_\_* features when missingness $>$ 0.80 is too aggressive given the extreme sparsity---especially when anchoring is fragile. It causes a feedback loop: fewer anchored patients $\rightarrow$ higher missingness $\rightarrow$ more drops $\rightarrow$ less clinical signal.

Change

* Switch from missing-rate thresholds to a value-based shortlist for longitudinal features:

  * Keep for key variables (DAS28, CRP, ESR, HAQ): last, days\_to\_start\_last, measured\_any, maybe max (CRP/ESR), and one windowed mean if enough measurements.

  * Drop high-variance low-support features (std in 0--90d) unless measurement count $\geq$2 and coverage is adequate.

* Gate window statistics by n\_measured (e.g., set std to NaN unless $\geq$2) to reduce noise.

Why

This preserves the few longitudinal summaries that can generalize and avoids the ``everything missing $\rightarrow$ everything pruned'' collapse.

---

6. Reduce dominance of treatment-choice \& socioeconomic proxies (improves generalization, often helps val AUROC)

Current SHAP shows AntiTNF\_or\_JAK\_First, SE, Scolar dominate. That can happen when clinical disease activity features are missing, but it can also reflect proxy learning that doesn't validate well.

Change (preprocessing-level)

* Once anchoring is fixed, ensure disease activity/inflammation features are reliably emitted (baseline + last-pre + recency).

* Consider coarsening high-cardinality socioeconomic variables (merge rare levels more aggressively than MIN\_CAT\_FREQ=5, or map to ordered bins if truly ordinal) to reduce overfitting.

Why

Shifts model capacity toward clinically causal-ish predictors (baseline disease state), which typically validates better.

---

7. Add diagnostics that directly optimize for ``usable longitudinal coverage''

To prevent repeating Step7's failure mode, add metadata counters like:

* \%patients with $\geq$1 pre-start row

* \%patients with baseline\_gap\_days within 0--30/31--90/91--365

* distribution of inferred start\_est spread (anchor uncertainty)

Why

These are leading indicators for downstream AUROC in this task; you can't improve the metric if preprocessing zeroes out the key signal again.

---

If you only change two things for the biggest expected AUROC gain: (1) relax/replace anchoring to recover pre-treatment rows for most patients, and (2) stop nulling meds/comorbidities in static-only mode.

\end{sysmsg}

\subsubsection{MD Evaluator Output (DoctorAgents-DS)}
\label{example:mt_eval}
\leavevmode\par
\begin{sysmsg}[MD Evaluation]

PART 1 --- History Analysis

* Performance volatility + collapse events indicate data/feature instability more than model choice.

  * Step 3 ``all-zero SHAP / AUROC 0.5'' strongly suggests degenerate features (near-constant after preprocessing, or leakage/QA features removed leaving almost no signal) and/or broken scaling/selection (e.g., extremely small C).

  * Step 4 ``train 1.0 / val 0.238'' is classic high-leakage / high-variance behavior (tiny n=58, many features, plus time-alignment features that can accidentally encode outcomes/care pathway).

* Anchoring strategy is the dominant failure mode.

  * Step 7 strict anchoring produced 55/58 ``anchor missing / no prestart / static-only fallback'', so the model learned mostly socioeconomic / treatment-choice proxies (SHAP: Scolar/Emploi/SE, AntiTNF\_or\_JAK\_First), not clinical trajectory.

  * Step 8 swings to anchor success 100\%, which is good for coverage but risky: removing plausibility gates can reintroduce post-start contamination unless pre-start identification is rock solid.

* Feature engineering sparsity is biting hard.

  * Step 8 metadata shows multiple \_\_std\_0\_90d features with missing rate 1.0 (across all patients). That's either:

    * a bug in window extraction (no rows fall in 0--90), or

    * nearly everyone has $<$2 visits in that window so std is undefined, and you're emitting all-NaN columns.

PART 2 --- Training Pipeline Correctness and Quality

* Pipeline is ``correct'' mechanically, but it is likely misaligned with the preprocessing output.

  * Your preprocessing already does: pruning, median impute, clipping, RobustScaler, OHE, missing indicators.

  * Training code imputes again, removes variance again, and for LR StandardScales again. Double transforms can:

    * distort the meaning of already-robust-scaled variables,

    * change the effective regularization strength,

    * and amplify noise for tiny datasets.

* Potential hard bug masked by current data export: if any column were truly all-NaN, SimpleImputer(median) would normally error. Since it doesn't, it suggests the exported CSV is already fully imputed---making the training-time imputer unnecessary at best.

* Model selection uses validation AUROC to pick the winner after CV shortlisting. With n=58, this can overfit to the single validation split; it's not ``wrong'' for a leaderboard-style val metric, but it will be unstable and can reward leakage.

PART 3 --- Model Selection, Hyperparameters, and Training Behavior

* p $>>$ n regime (371 features vs 58 patients): univariate SelectKBest (k 10--30) is doing most of the work. This is high-variance and very sensitive to tiny shifts in preprocessing/anchoring.

* HGB + f\_classif feature selection is a questionable pairing.

  * Trees don't need scaling; they also don't benefit much from a linear univariate filter that can discard interaction features.

* Overfitting risk remains high even when train AUROC isn't 1.0, because:

  * selection is unstable (small k, many correlated engineered features),

  * time-alignment features can act as proxies,

  * and class\_weight/sample\_weight balancing can increase variance further in tiny samples.

PART 4 --- Improvement Strategy \& Next-Step Recommendations

A) Fix anchoring/window feature reliability first (highest ROI)

1. Add sanity gates back, but softer than Step 7.

   Keep broad coverage, but reject anchors that imply impossible timelines.

   * Pseudocode idea:

     * compute anchor candidates; choose one only if it yields a reasonable fraction of visits with days\_to\_start $\geq$ 0 and within a max horizon (e.g., 0--730), and if the implied start is not after the earliest visit by a large margin.

2. Stop emitting ``always missing'' engineered features.

   * If n\_measured $<$ 2 in a window, either:

     * don't create std at all, or

     * set std = 0 and rely on n\_measured to indicate reliability (preferable to all-NaN columns that become artifacts after imputation).

3. Revisit the 0--90 / 91--365 split.

   If most patients have sparse visits, those windows are too granular. Consider:

   * a single 0--365 window, or

   * 0--180 and 181--365, or

   * adaptive windows based on available visits (but be careful to keep it time-safe).

B) Reduce care-pattern leakage

* In step 6/7/8 you still have measurement density signals (n\_measured, recency, sometimes visit counts). These can dominate. Try:

  * removing raw visit-count features,

  * or restricting them to a single coarse ``has\_any\_measurement'' reliability flag per domain.

C) Align preprocessing with modeling (avoid double transforms)

* Either:

  * export minimally processed features (no scaling/imputation) and keep all transforms in the sklearn Pipeline, or

  * export fully model-ready features and remove imputer/scaler from the training pipeline.

* Also consider producing two feature matrices: one scaled for linear models and one unscaled for trees, to avoid hurting HGB with clipping/scaling choices optimized for LR.

D) If val AUROC doesn't improve in $\sim$3 iterations: change model family/selection

Given tiny n, high-dim, and instability:

* Prefer a single strongly-regularized linear model (ridge / elastic net) using all reasonably filtered features, and drop SelectKBest (or set k much higher, e.g., 80--200) to reduce selection variance.

* Alternatively, try Bayesian/empirical Bayes logistic regression (conceptually) or at least stronger priors via smaller C with stability selection---because your main issue is variance, not bias.

Quick ``next run'' checklist

* Verify (per split) distribution of n\_visits\_prestart\_0\_90, n\_visits\_prestart\_91\_365, and the fraction with n\_measured$\geq$2 for key PROs; if near-zero, remove std features and/or change windows.

* Confirm that ``pre-treatment rows'' are truly pre-treatment: audit a few patients' inferred start vs actual visit dates.

* Run SHAP on a model where clinical activity features are present; if SHAP is again dominated by socioeconomic/treatment-choice, anchoring/feature availability is still failing.

\end{sysmsg}

\subsubsection{MD Backpropagation Output (DoctorAgents-DS)}
\label{example:mt_fb}
\leavevmode\par
\begin{sysmsg}[MD Feedback]

Highest-impact issues in the current training code (hurting AUROC)

* You're ``preprocessing twice'' (and inconsistently): the exported CSV is already heavily processed (imputation/clipping/RobustScaler/OHE/missing flags per metadata), yet the training pipeline applies median imputation + variance filtering + StandardScaler + SelectKBest again. This can (a) distort already-robust-scaled features, (b) change the effective regularization strength, and (c) amplify noise in a tiny-n regime.

* Univariate SelectKBest(f\_classif) is extremely high-variance at n=58, p=371 and is likely selecting unstable proxies (as seen when anchors were missing, where socioeconomic/treatment-choice dominated SHAP). Even when anchors are fine, k=10--30 is so small that tiny perturbations swap the chosen set.

* HGB + SelectKBest(f\_classif) is a mismatched combo: trees don't need scaling, and a linear univariate filter can delete interaction/threshold signal that trees would otherwise use. If HGB is worth trying here, it should see a broader, less pre-filtered feature set (or a tree-appropriate selector).

* CV robustness is brittle given class counts: fixed n\_splits=5 can still yield folds with too few positives/negatives depending on imbalance; your logic then skips folds and may invalidate otherwise good configs (cv\_score=-1e9). This can silently bias the search toward ``configs that happen not to break'' rather than best AUROC.

---

Concrete changes to the code that are likely to improve validation AUROC

1. Make the modeling pipeline consistent with what preprocessing exports (choose one owner of transforms).

   * If the CSV is truly ``model-ready'' (already imputed/scaled), then drop training-time SimpleImputer and StandardScaler (at minimum for LR) and keep only minimal guards (e.g., constant-column removal).

   * If you want transforms in-model (preferred), then export less processed data and let the sklearn Pipeline do impute/scale/OHE---right now you're in an unstable middle.

2. Add train-time dropping of ``structurally missing'' engineered columns (especially window stats).

   * Your metadata shows many window \_\_std\_0\_90d are missing rate 1.0. Even if imputation makes them constants, they waste degrees of freedom and add selector noise.

   * Add a simple filter before model search: drop columns with missing-rate $\geq$ (e.g.) 0.98 or with ``unique after impute'' $\leq$ 1. This directly reduces SelectKBest variance.

3. Replace or de-emphasize SelectKBest (or at least make it much less aggressive).

   * Try no KBest for LR and rely on ridge/elasticnet regularization (much more stable in p$>>$n).

   * If you keep KBest, push k much higher (think 80--250) and/or use stability selection: compute selection frequency across CV folds and keep features that recur (this often beats picking a single brittle top-k).

4. Expand the LR regularization grid (current grid likely misses good regimes).

   * Your step history already found useful solutions at C$\approx$3 (Step 6), but Step 7's search tops out at C=0.3 for L2 and elastic. That's an unnecessary ceiling.

   * Also consider solver choices tailored for tiny data (e.g., liblinear for L1/L2) to reduce convergence quirks vs saga in small n.

5. Fix HGB to be tree-native (if you keep it).

   * Let HGB consume more features: remove SelectKBest (or use a very high k) and remove StandardScaler (not used now, good).

   * Consider using HGB's native NaN handling by not imputing for HGB at all (your current SimpleImputer erases informative missingness patterns that can matter in longitudinal ``measured vs not measured'' data).

   * Tune parameters that actually control overfit in small n: max\_leaf\_nodes, min\_samples\_leaf, max\_depth, learning\_rate, l2\_regularization. Right now the HGB grid is narrow and may not find a good bias/variance point.

6. Make CV split settings adaptive to class counts to avoid fold skipping and invalidation.

   * Set n\_splits = min(5, min(n\_pos, n\_neg)) (or even 3 if minority class is tiny), and/or switch to repeated stratified shuffle split for scoring robustness.

   * Also: clone the pipeline inside each fold (sklearn.base.clone) instead of reusing the same instance; this avoids subtle state carryover issues and makes results more reproducible.

7. Use ensembling to reduce variance (very effective at n=58).

   * Instead of selecting a single `best'' spec, average predicted probabilities from the top-N diverse specs (e.g., ridge LR + elastic LR + HGB) weighted by CV score. This often improves AUROC in exactly this `small sample, unstable feature selection'' regime.

8. Guard against proxy-dominance as a model-selection criterion (to improve hidden/generalization AUROC).

   * Add an optional `proxy blacklist'' experiment (education/employment/treatment-choice columns) and treat it as a separate candidate family; select between `full'' vs ``no-proxy'' by CV. This helps when anchoring/window coverage is imperfect and the model otherwise latches onto non-clinical correlates that don't generalize.

---

One ``tell'' to act on immediately

If Step 8 still has many columns with missing rate 1.0 (or near-constant after impute), your current SelectKBest(k=10--30) is effectively doing feature roulette. Any change that (a) removes those columns and (b) reduces reliance on aggressive univariate selection is very likely to move AUROC more than tweaking HGB depth or learning rate.

\end{sysmsg}

\subsubsection{Logger Output}
\label{example:log}
\leavevmode\par
\begin{sysmsg}[Note Log of \model-DS after Summarization (SUMMA)]
Step 0$\rightarrow$1: Major feature reduction; start inferred from earliest DReelle + median baseline\_time\_to\_DateDeb; baseline latest eligible within 365d; rare categories$\rightarrow$Other; all-numeric clipping; train-schema reindex; CV-ranked/val-shortlisted LR/HGB with weights/early stopping. Perf: GradientBoostingClassifier train/val AUROC 1.0/0.69 (Step 0) $\rightarrow$ best LogisticRegression\_l2 train/val 1.0/0.571 (Step 1).\\

Step 1$\rightarrow$2: Enforced baseline\_time\_to\_DateDeb-based start (median per-row DateDeb\_i); baseline closest pre-start (+ outside-365d flag); separated strict 365d summaries from baseline/serology max; added timing/consistency flags + comorbidity burden; label-consistency collapse + flag; refined clipping/scaling (exclude binary/counts; data-driven missing indicators); leakage-reduced Pipeline LR(L2)+HGB with fold-wise weighting and penalized CV selection. Perf unchanged: train 1.0, val 0.571.\\

Step 2$\rightarrow$3: Two-stage start inference with start\_method; removed leakage/QA flag features in favor of recency; simplified longitudinal markers; curated missing indicators; broadened LR(ElasticNet/L2)+expanded HGB search with optional blending. Features 199$\rightarrow$111 (OHE 43$\rightarrow$19). Perf collapse: best LR\_elasticnet C=1e-5, train/val 0.5/0.5; SHAP all-zero.\\

Step 3$\rightarrow$4: Added start\_confidence + anti-contamination gate (longitudinal features forced missing/0 when start\_confidence$<$2); baseline closest pre-start with baseline\_stale/baseline\_gap\_bin; trajectory features (last/delta/slope\_per\_day; CRP/ESR log1p); binned visit intensity; moved meds/comorbidities/serology and many counts to categorical/OHE; missing indicators for numeric $>$5\% missingness; standardized prune+median-impute; always-balanced weighting (class\_weight="balanced" + sample\_weight); stricter CV penalties/degeneracy filtering; no ensembling + fallback guardrail. Perf: best LR\_l2 C=1.0, train/val 1.0/0.238 (instability/overfitting signal).\\

Step 4$\rightarrow$5: Dropped confidence-gated date reconstruction; used baseline\_time\_to\_DateDeb as sole anchor with always-emitted baseline snapshots + sparse window summaries; meds/comorbidities numeric with pre-scaling numeric pruning; k-best feature-selected LR with revised weighting/CV/selection and SHAP mapping. Perf improvement: LR\_kbest\_ridge k=15 val AUROC 0.619 (from 0.238).\\

Step 5$\rightarrow$6: Reconstructed robust anchor btt\_used with anchor\_mode + baseline\_gap\_bin; broader 365d coverage; added visit/trend/mean features and `time-safe'' last-prestart fallback values; stronger category canonicalization; modeling simplified to LR/HGB `views'' with hard no-NaN inputs and StratifiedShuffleSplit scoring. Perf regression: 0.819/0.619 $\rightarrow$ 0.983/0.524 (lr\_elastic).\\

Step 6$\rightarrow$7: Moved from flexible anchoring with single 0--365d window and manual-view LR/HGB (StratifiedShuffleSplit; train/val 0.983/0.524) to strict pre-treatment anchoring with static-only fallback and two windows (0--90/91--365d; mean/std/max/log1p; latest-pre serology + optional quantile bins), engineered-feature drop $>$0.80, explicit QA (anchor missing/no-prestart/static-only for 55/58), and sklearn Pipelines (imputation/near-constant removal/scaling/SelectKBest) tuned under RepeatedStratifiedKFold. Perf: train/val 0.893/0.571.\\

Step 7$\rightarrow$8: Switched to robust-consensus anchor without [0,365] gate (dt\_plus\_btt for 58/58; anchor\_n\_support/anchor\_iqr\_days); static covariates from latest row regardless of pre-start; meds/comorbidities ternary categoricals even when anchoring fails; missingness drop relaxed to $>$0.95; categorical missing = ``Missing''; serology qbins categorical one-hot; modeling changed to direct array training after one-time filtering with StratifiedShuffleSplit and simpler penalties. Perf jump: val AUROC 0.571 (lr\_l2 C=0.3,k=15) $\rightarrow$ 0.905 (lr\_elastic\_saga C=0.1,l1\_ratio=0.9); train/val 0.867/0.905.\\

Step 8$\rightarrow$9: Changed to earliest-3-visit, prestart-only, monotonic-consistency--scored anchor (anchor\_dt\_btt\_corr) removing fallback leakage; used decayed means, stricter std rules, added deltas/support gates; tighter pruning (missingness$>$0.90), higher missing-indicator threshold; serology \_\_measured/\_\_positive\_proxy; broader CV with overfit/degeneracy filtering and no HGB early stopping. Perf drop: LR elastic C=0.01,l1\_ratio=0.1, train/val 0.588/0.738 (from 0.867/0.905).\\

Step 9$\rightarrow$10: Switched to correlation-gated/reconstructed anchoring with validity/shifted-late suppression of window features; dual baseline snapshots+deltas; invariant-any-row categoricals and any-row latest serology; looser/always-keep pruning; sparse-friendly window summaries; single validation-split search; selected early-stopped HGB. Perf: train/val 0.8/0.595.\\

Step 10$\rightarrow$11: Moved to two-candidate ($\pm$) median start estimator with anchor uncertainty/availability flags; broader baseline/recency indicators; 0--180/181--365d windows with NaN/window\_available\_*; added 0--365d aggregates; tighter leakage/invariant handling and stricter pruning; CV-penalized model screening with pipeline imputation/feature filtering. Perf regression: val AUROC 0.595 (HGB) $\rightarrow$ 0.464 (LR L1 saga C=0.003).

\end{sysmsg}

\subsection{Examples of code snippet generated by \model-DS}
\label{apsec:example_code}

This section presents examples of generated code snippets emphasizing the feature-engineering capabilities of \model-DS.

\subsubsection{Mortality Prediction}
\label{code:mort_example_code}
\leavevmode\par
\begin{sysmsg}["bicarbonate\_blood\_delta\_per\_hr" (L63-66 and L353-361)]
\begin{lstlisting}
TRAJ_LABS = [
    "creatinine_blood", "urea_nitrogen_blood", "lactate_blood", "bicarbonate_blood",
    ...
]

......

if c in TRAJ_LABS:
    delta = pd.to_numeric(last_s, errors="coerce") - pd.to_numeric(first_s, errors="coerce")
    elapsed_hr = (last_t - first_t).dt.total_seconds() / 3600.0
    dph = safe_div(delta, elapsed_hr.replace(0.0, np.nan))
    out[f"{c}_first"] = first_s
    out[f"{c}_delta_last_minus_first"] = delta.where(n_meas >= 2.0, np.nan).astype("float32")
    out[f"{c}_delta_per_hr"] = dph.where(n_meas >= 2.0, np.nan).astype("float32")
    out[f"{c}_trend_defined"] = (n_meas >= 2.0).astype("int8")
\end{lstlisting}
\end{sysmsg}

\begin{sysmsg}["gcs\_eye\_last" (L35-38 and L283-294)]
\begin{lstlisting}
VITALS = [
    ...
    "gcs_total", "gcs_eye", "gcs_motor", "gcs_verbal", "rass"
]

......

val_nonnull = dfw.loc[dfw[c].notna(), [ID_COL, c, EVENT_TIME_COL]]
if len(val_nonnull):
    first_s = val_nonnull.groupby(ID_COL, sort=False)[c].first().reindex(subjects)
    last_s = val_nonnull.groupby(ID_COL, sort=False)[c].last().reindex(subjects)
    first_t = val_nonnull.groupby(ID_COL, sort=False)[EVENT_TIME_COL].min().reindex(subjects)
    last_t = val_nonnull.groupby(ID_COL, sort=False)[EVENT_TIME_COL].max().reindex(subjects)
...
out[f"{c}_last"] = last_s
\end{lstlisting}
\end{sysmsg}

\begin{sysmsg}[{
\parbox[t]{0.95\linewidth}{
"\nolinkurl{mean_arterial_pressure_mean_24_48h}" 
(L75, L270-272, and L307-315)
}
}]
\begin{lstlisting}
BIN_VITALS = ["mean_arterial_pressure", "heart_rate", "respiratory_rate", "spo2", "fio2", "temperature"]

......

dt_from_start = (dfw[EVENT_TIME_COL] - dfw[T0_COL]).dt.total_seconds() / 3600.0
recent = (dt_from_start >= 0.0) & (dt_from_start <= 24.0)
early = (dt_from_start > 24.0) & (dt_from_start <= 48.0)

......

for c in BIN_VITALS:
    if c in dfw.columns:
        g_recent = dfw.loc[recent].groupby(ID_COL, sort=False)[c].mean().reindex(subjects)
        g_early = dfw.loc[early].groupby(ID_COL, sort=False)[c].mean().reindex(subjects)
        ...
        out[f"{c}_mean_0_24h"] = g_recent
        out[f"{c}_mean_24_48h"] = g_early
\end{lstlisting}
\end{sysmsg}

\begin{sysmsg}["n\_glucose\_blood\_meas\_48h" (L41-45 and L265-268)]
\begin{lstlisting}[language=Python]
LABS = [
    ...
    "urea_nitrogen_blood", "creatinine_blood", "glucose_blood", "calcium_total_blood",
    ...
]
......

count_vars = sorted(set([c for c in (vitals_present + labs_present) if c in (set(VITALS) | set(TRAJ_LABS) | set(LAB_RECENCY) | set(VITAL_TRENDS) | set(BIN_VITALS) | set(BIN_LABS) | {"gcs_total","gcs_eye","gcs_motor","gcs_verbal"})]))
for c in count_vars:
    out[f"n_{c}_meas_48h"] = dfw[c].notna().groupby(dfw[ID_COL], sort=False).sum(min_count=1).reindex(subjects).fillna(0.0).astype("float32")
    out[f"{c}_measured_any_48h"] = (out[f"n_{c}_meas_48h"] > 0.0).astype("int8")
\end{lstlisting}
\end{sysmsg}

\subsubsection{Readmission Prediction}
\label{code:readm_example_code}
\leavevmode\par
\begin{sysmsg}[{
\parbox[t]{0.95\linewidth}{
"heart\_rate\_count\_log1p\_last6h" (L69, L529-535, and L763-775)
}
}]
\begin{lstlisting}
CORE_VITALS_6H = set(["heart_rate", "respiratory_rate", "systolic_bp", "mean_arterial_pressure", "spo2", "fio2", "temperature"])

......

if concept in core_vitals_6h_keep:
    w2 = float(vital_small_h)
    last2, rec2, first2, first_rec2 = _first_last_in_window(ev, src_col, concept, w2)
    cnt2 = _count_in_window(ev, src_col, concept, w2)
    dyn[f"{concept}__last_in_last{int(w2)}h"] = last2
    dyn[f"{concept}__last_in_last{int(w2)}h_hours_before_discharge"] = rec2
    dyn[f"{concept}__count_last{int(w2)}h"] = cnt2

......

for c in cnt_cols:
    base, win = _parse(c)
    ...
    if base in curated_concepts:
        out[f"{base}__measured_any_last{win}h"] = (s > 0).astype("int8")
        cap = float(labs_cap)
        if base in VITAL_RANGES:
            cap = float(vitals_caps[1] if win == vitals_caps[0] else vitals_caps[3])
        out[f"{base}__count_capped_last{win}h"] = np.minimum(s, float(cap)).astype("float32")
        out[f"{base}__count_log1p_last{win}h"] = np.log1p(s).astype("float32")
        kept += 1
\end{lstlisting}
\end{sysmsg}

\begin{sysmsg}[{
\parbox[t]{0.95\linewidth}{"respiratory\_rate\_slope\_last6h\_per\_hr" (L69, L505, and L529-553)}
}]
\begin{lstlisting}
CORE_VITALS_6H = set(["heart_rate", "respiratory_rate", "systolic_bp", "mean_arterial_pressure", "spo2", "fio2", "temperature"])

......

vital_slope_keep_6h = set(["heart_rate", "mean_arterial_pressure", "systolic_bp", "respiratory_rate", "spo2", "fio2", "temperature"])

......

if concept in core_vitals_6h_keep:
    w2 = float(vital_small_h)
    last2, rec2, first2, first_rec2 = _first_last_in_window(ev, src_col, concept, w2)
    cnt2 = _count_in_window(ev, src_col, concept, w2)
    ...
    if (concept in vital_slope_keep_6h) and (not last2.empty) and (not first2.empty):
        delta2 = (last2 - first2).replace([np.inf, -np.inf], np.nan).astype("float32")
        span2 = (first_rec2 - rec2).replace([np.inf, -np.inf], np.nan).astype("float32")
        span2 = span2.where(cnt2 >= 2, np.nan)
        slope2 = (delta2 / span2.replace(0, np.nan)).replace([np.inf, -np.inf], np.nan).astype("float32")
        dyn[f"{concept}__slope_last{int(w2)}h_per_hr"] = slope2
        dyn[f"{concept}__n_ge2_last{int(w2)}h"] = (cnt2.reindex(dyn.index).fillna(0).astype("float32") >= 2).astype("int8")
\end{lstlisting}
\end{sysmsg}




\subsubsection{LOS Prediction}
\label{code:los_example_code}
\leavevmode\par
\begin{sysmsg}[{
\parbox[t]{0.95\linewidth}{
"\nolinkurl{fio2 / po2\_blood"\_mean\_last6h, and "lactate\_blood\_recency\_hours}" (L1120-1124, L257-281, and L492-516)}
}]
\begin{lstlisting}
workflow_keys_small = [c for c in [
    "fio2", "spo2", "gcs_total", "gcs_verbal", "gcs_motor",
    "lactate_blood", "ph_blood", "po2_blood", "pco2_blood",
    "chloride_blood", "rdw_blood", "ptt_blood", "base_excess_blood", "creatinine_blood"
] if c in train_raw.columns]

......

def first_last6h_means_counts(dfv: pd.DataFrame, col: str):
    ...
    last_start = tmp["_pred"] - pd.to_timedelta(6.0, unit="h")
    last_end = tmp["_pred"]
    ...
    s_last = last_blk.groupby(ID_COL, sort=False)[col].mean().astype("float64") if not last_blk.empty else pd.Series(dtype="float64")
    ...

......

workflow_keys_small = [c for c in workflow_keys_small if c in dfv.columns]
...
for c in workflow_keys_small:
    ...
    s_first, s_last, c_first, c_last = first_last6h_means_counts(dfv, c)
    ...
    wf = pd.DataFrame({
        ...
        f"{c}__mean_last6h": s_last_r,
        ...
    }, index=cnt.index)
\end{lstlisting}
\end{sysmsg}

\begin{sysmsg}[
\text{"spo2\_fio2\_ratio" (L332-400, L598-604, and L744-753)}
]
\begin{lstlisting}
def paired_ratio_aggregates(dfv: pd.DataFrame, numer_col: str, denom_col: str, prefix: str, max_gap_hours: float = 2.0):
    ...
    ratio = (g[numer_col] / denom_ff).where(ok, np.nan).astype("float64")
    ...
    out[f"{prefix}__paired_mean_last6h__defined"] = res.apply(lambda x: float(x[3]) if isinstance(x, tuple) else 0.0).astype("float64")

......

pr_parts = []
if "spo2" in dfv.columns and "fio2" in dfv.columns:
    pr_parts.append(paired_ratio_aggregates(dfv, "spo2", "fio2", "sf_ratio", float(PAIR_MAX_GAP_HOURS)))

......

elif has("spo2__last") and has("fio2__last"):
    denom = fnum(out["fio2__last"]).replace(0.0, np.nan)
    out["spo2_fio2_ratio"] = fnum(out["spo2__last"]) / denom
    out["spo2_fio2_ratio__defined"] = ((fnum(out.get("spo2__measured", np.nan)).fillna(0.0) > 0) & (fnum(out.get("fio2__measured", np.nan)).fillna(0.0) > 0)).astype("float64")
\end{lstlisting}
\end{sysmsg}

\subsubsection{RA Anti-TNF Response Prediction}
\label{code:ra_example_code}
\leavevmode\par
\begin{sysmsg}[{
\parbox[t]{0.95\linewidth}{\nolinkurl{"SE\_2.0", "GrFonct\_2.0", and "AntiTNF\_or\_JAK\_First\_3"} (L131, L260-263, and L299-323)
}
}]
\begin{lstlisting}
cats = ["Sexe", "Scolar", "SE", "Emploi", "StatFum", "StatMar", "ruralurban", "AntiTNF_or_JAK_First", "ACR2010", "Activity", "GrFonct"]

......

for c in cats:
    if c in g.columns:
        v = base.get(c, np.nan)
        row[c] = str(v) if pd.notna(v) else "Unknown"

......

cat_cols = [c for c in train_pat.columns if c not in protected and train_pat[c].dtype == "object"]
...
enc = OneHotEncoder(handle_unknown="ignore", sparse_output=False, dtype=np.float32)
enc.fit(train_cat.values)
ohe_names = list(enc.get_feature_names_out(cat_cols))
\end{lstlisting}
\end{sysmsg}




\begin{sysmsg}["PGA\_last\_pre" and "EGA\_last\_pre" (L124-L128 and L244-L247)]
\begin{lstlisting}
baseline_like_numeric = [
    "DAS28CRP_4", "Crp", "Esr", "PGA", "EGA", "M_HAQ", "Raideur", "Fatigue", "Sommeil",
    "FscGB", "FscHb", "FscPLQ", "FscPMN", "IMC", "Sharp_total",
    "Duree_symp_mois", "Age_V1", "time_first_symptom_to_DateDeb"
]

......

for c in baseline_like_numeric:
    if c in g.columns:
        v = _as_numeric(base.get(c, np.nan))
        row[f"{c}__last_pre"] = float(v) if pd.notna(v) else np.nan
\end{lstlisting}
\end{sysmsg}

\begin{sysmsg}[\text{"FscPLQ\_delta\_365d" (L134, L71-L117, and L265-L266)}]
\begin{lstlisting}
window_vars = ["DAS28CRP_4", "Crp", "Esr", "PGA", "EGA", "M_HAQ", "Raideur", "Fatigue", "Sommeil", "FscPLQ"]

......

def _summarize_window(elig, tcol, vars_list):
    ...
    for c in vars_list:
        ...
        first_v = x.loc[first_i]
        last_v = x.loc[last_i]
        res[f"{c}__delta_365d"] = float(last_v - first_v) if (pd.notna(first_v) and pd.notna(last_v)) else np.nan
    return res

......

win_summ = _summarize_window(elig_365 if has_365 else elig_365, "_t", window_vars)
row.update(win_summ)
\end{lstlisting}
\end{sysmsg}


\end{document}